\documentclass[11pt]{article}
\usepackage{amssymb}
\usepackage{amsfonts}

\usepackage[final]{acl}

\usepackage{times}
\usepackage{latexsym}

\usepackage[T1]{fontenc}

\usepackage[utf8]{inputenc}

\usepackage{tcolorbox}
\usepackage{amsmath}
\tcbuselibrary{breakable}

\usepackage{microtype}

\usepackage{inconsolata}

\usepackage{graphicx}

\usepackage{tabularx}
\usepackage{pifont}
\usepackage[dvipsnames]{xcolor}
\definecolor{darkgreen}{RGB}{0,120,0}

\title{Whose Facts Win? LLM Source Preferences under Knowledge Conflicts}

\author{Jakob Schuster \\
  Heidelberg University \\
  \texttt{schuster@cl.uni-heidelberg.de} \\ \And
  Vagrant Gautam \\
  Heidelberg Institute for Theoretical Studies \\
  \texttt{vagrant.gautam@h-its.org} \\ \AND
  Katja Markert \\
  Heidelberg University \\
  \texttt{markert@cl.uni-heidelberg.de} \\}

\begin{document}
\maketitle
\begin{abstract}
As large language models (LLMs) are more frequently used in retrieval-augmented generation pipelines, it is increasingly relevant to study their behavior under knowledge conflicts.
Thus far, the role of the \textit{source} of the retrieved information has gone unexamined.
We address this gap with a novel framework to investigate how source preferences affect LLM resolution of inter-context knowledge conflicts in English, motivated by interdisciplinary research on credibility.
By using synthetic sources, we study preferences for different types of sources without inheriting the biases of specific real-world sources.
With a comprehensive, tightly-controlled evaluation of $13$ open-weight LLMs, we find that LLMs prefer institutionally-corroborated information (e.g., government or newspaper sources) over information from people and social media.
However, these source preferences can be reversed by simply repeating information from less credible sources.
To mitigate repetition effects and maintain consistent preferences, we propose a novel method that reduces repetition bias by up to $79.2$\%, while also maintaining at least $72.5$\% of original preferences.
We release all data and code to encourage future work on credibility and source preferences in knowledge-intensive NLP.\footnote{\url{https://github.com/JaSchuste/llm-source-preference}}
\end{abstract}

\section{Introduction}

Since their rapid adoption as conversational assistants \citep{ouyang2022training}, large language models (LLMs) are now widely used for knowledge-intensive tasks such as question answering, summarization, and information retrieval \citep{shah-bender-2024-envisioning}.
However, when forced to rely on parametric knowledge encoded during pre-training, LLMs often fabricate factually incorrect statements \citep{ji2023survey}.
To reduce such errors, they are commonly embedded in retrieval-augmented generation (RAG) pipelines to ground generation in evidence from external sources \citep{lewis2020retrieval}.

\begin{figure}
    \centering
    \includegraphics[width=\columnwidth]{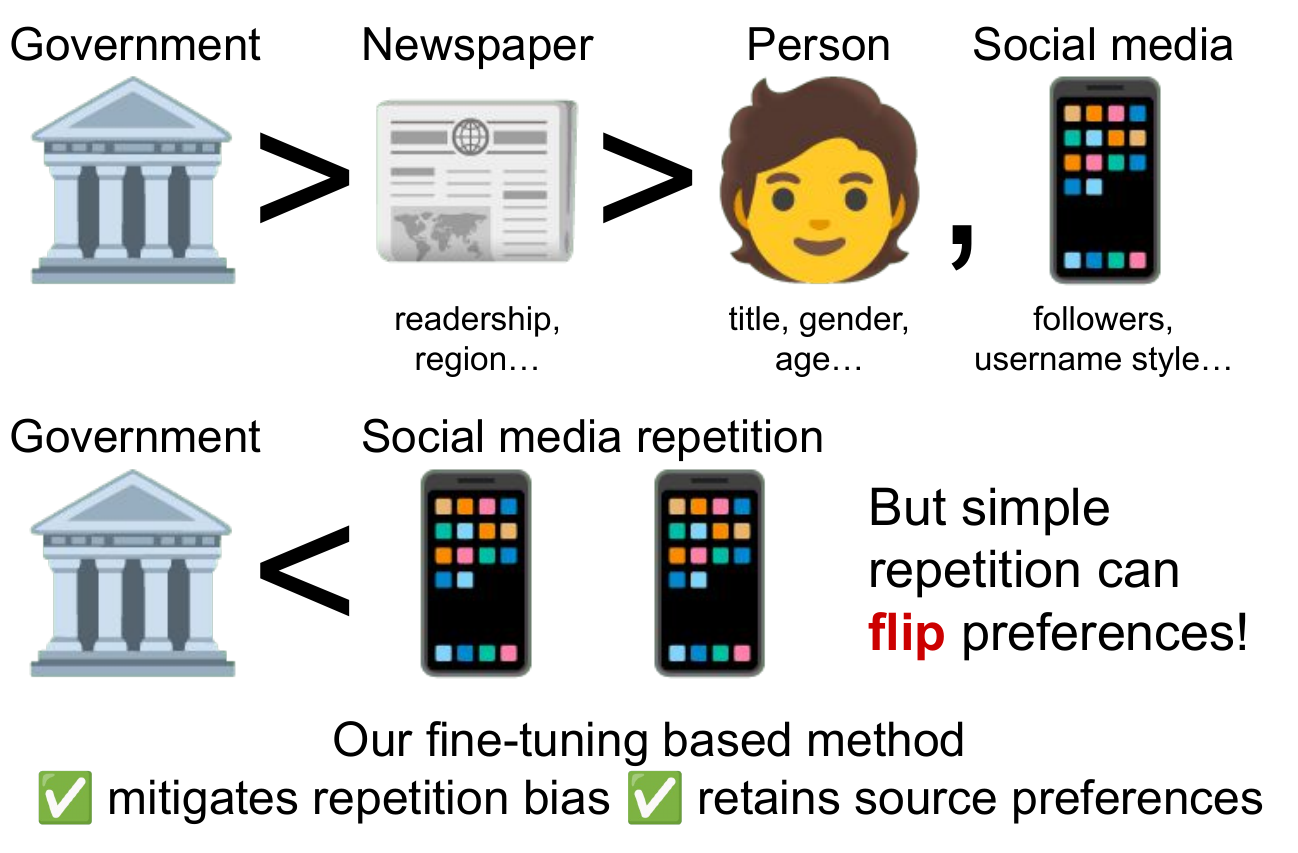}
    \includegraphics[width=\columnwidth]{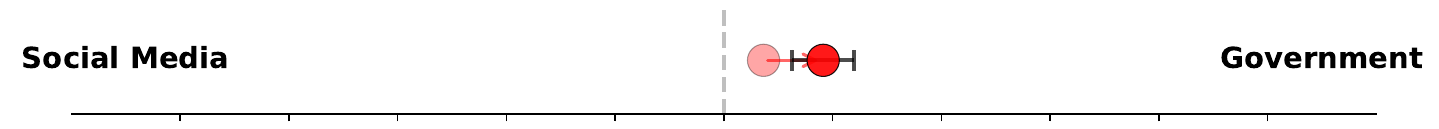}
    \includegraphics[width=\columnwidth]{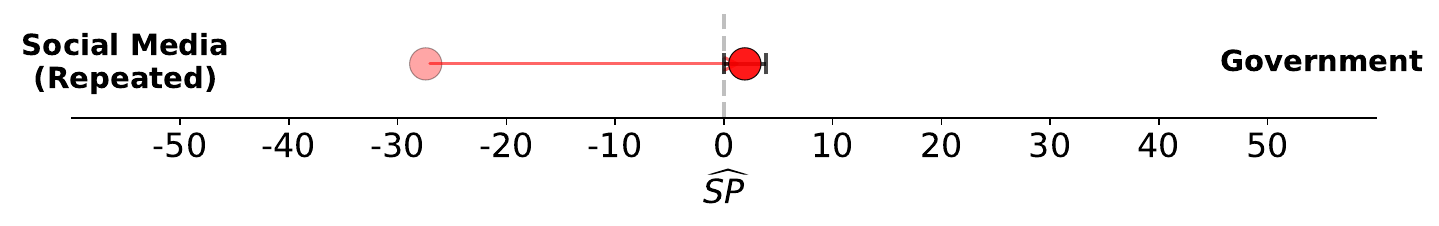}
    \caption{Evaluating $13$ LLMs on source and knowledge conflicts induces a consistent source credibility hierarchy. However, repeating information can flip preferences. Our method (darker) mitigates repetition bias and maintains original (lighter) source preferences.}
    \label{fig:one}
\end{figure}

While retrieval can ground answers in concrete evidence, it can also create knowledge conflicts between contexts, due to ambiguous named entities, outdated documents, or explicitly false or misleading information \citep{xu2024knowledge, pan-etal-2023-attacking}.
Previous work on inter-context conflicts has shown models to prefer more relevant retrieved passages \cite{chen2022rich}, contexts aligned with parametric knowledge \cite{xie2024adaptive}, frequent information \cite{jin2024tug}, as well as LLM-generated information \cite{tan2024blinded}.
However, no study thus far examines \textit{the role of the information source} in how LLMs resolve such conflicts.

\begin{figure*}[t]
    \centering
    \includegraphics[width=\linewidth]{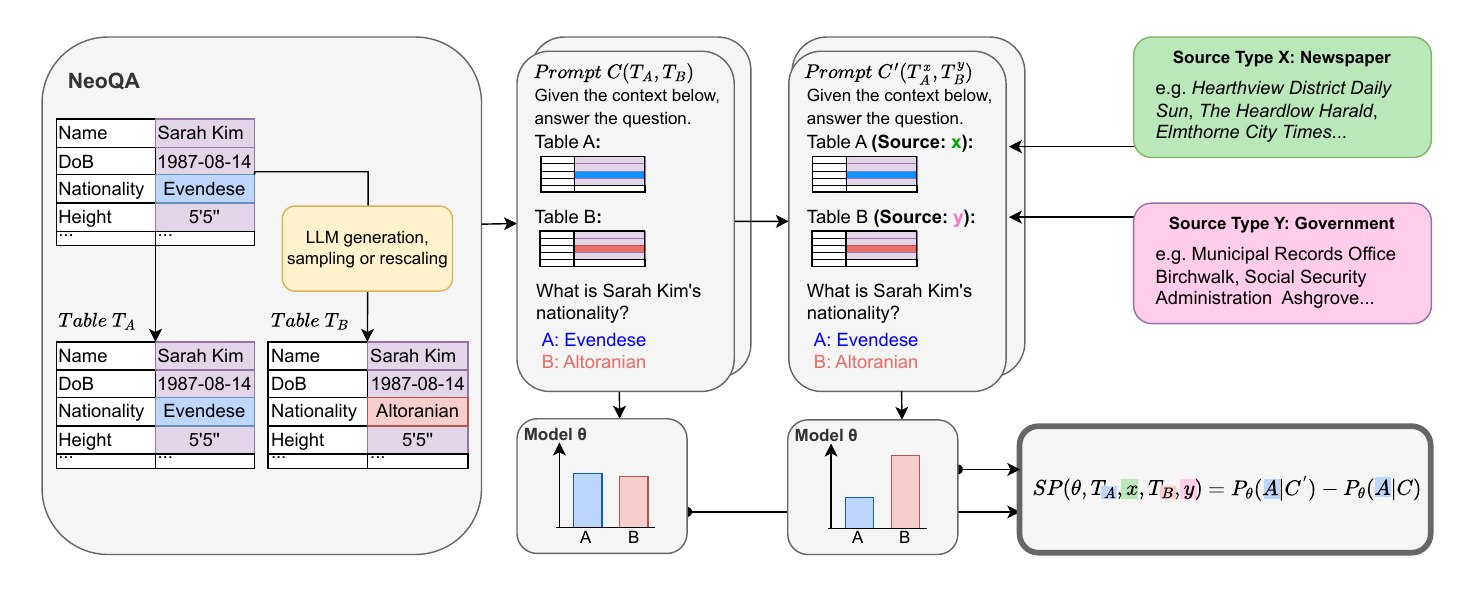}
    \vspace{-10mm}
    \caption{We measure the influence of source credibility on a model's output by observing how answer probabilities for conflicting information shift when attributed to a particular  source group.}
    \vspace{-2mm}
    \label{fig:setup}
\end{figure*}

We address this gap in the literature by investigating how LLMs resolve knowledge conflicts from different sources (e.g., government, newspaper, social media user, person) with various features (e.g., circulation of a newspaper, age of a person).
We do this by systematically evaluating $13$ models 
in a controlled multiple-choice question answering (MCQA) setting with synthetic conflicts and sources.
Synthetic conflicts allow us to isolate source preferences, ignoring the effects of pre-training knowledge.
Similarly, synthetic sources allow us to abstract to the Platonic ideal of different source types, ignoring biases related to the political leanings, readerships, etc., of real-world sources, and similar to human surveys on general institutional or media trust \citep{pewresearch2025how}.
Studying these abstracted preferences is necessary to understand how specific real-world scenarios conform to or deviate from them. 
Our central findings and contributions are:

\begin{itemize}
    \item With interdisciplinary grounding in credibility (\S\ref{sec:credibility}), we introduce a novel framework to study how source preferences affect LLM resolution of inter-context knowledge conflicts (\S\ref{sec:data-methodology}).
    \item Sources and their features significantly affect how LLMs resolve knowledge conflicts, and LLM conflict resolution follows a highly consistent \textit{source credibility hierarchy} (\S\ref{sec:source-preferences}).
    \item Repeating information from low-credibility sources can flip LLM source preferences (\S\ref{sec:repetition-bias}), showing a critical vulnerability of LLMs to disinformation as seen in \citet{NewsGuard_2024}.
    \item We propose a novel fine-tuning-based method which mitigates repetition bias by up to $79.2$\%, while also maintaining at least $72.5$\% of original source preferences, even for source types  not encountered in training (\S\ref{sec:mitigation}).
\end{itemize}

\noindent Our findings, summarized in Figure \ref{fig:one}, show that credibility and source preferences are rich aspects of research in RAG and QA, with important implications for a trustworthy information ecosystem.

\section{Background: Credibility}
\label{sec:credibility}

Credibility has a long history of examination in communication, psychology, cognitive sciences, media studies, and human-computer interaction \citep{Rieh_Danielson_2007}.
All key components of communication (source, message, medium, and recipient) are implicated in credibility judgements \citep{pornpitakpan-2004-persuasiveness}.
In this paper, however, we focus on judgments of \textit{source} credibility, i.e., attitudes towards the entity a message originates from \citep{hovland-weiss-1951-influence}.

Early research on source credibility asked people which version of a story they found most believable given conflicting reports from traditional print media, television, and radio sources \citep{hovland-weiss-1951-influence,Roper_1985}.
Later research began to disentangle multiple dimensions of source credibility \citep{whitehead-1968-factors,mccroskey-young-1981-ethos}, and to include the internet in research as a source and medium \citep{flanagin-metzger-2000}. 

In our research, rather than studying human credibility judgments, we focus on how source credibility affects \textit{LLM} decisions under knowledge and source conflicts.
We experiment with long-studied contrasts in source credibility, including newspapers, government, and social media.
Under \citeauthor{fogg-tseng-1999-elements}'s \citeyearpar{fogg-tseng-1999-elements} framework of credibility, we investigate \textit{presumed credibility} (general assumptions about a source's credibility) as well as \textit{reputed credibility} (judgments based on third-party reports).
By using equally plausible factual knowledge conflicts, we avoid variation in message credibility, allowing us to isolate source credibility in LLMs.

Studying credibility with LLMs presents several unique challenges not encountered in human studies of credibility:
We must control for position bias, token bias, and result stability, and we must design a framework to isolate source preferences at an abstract level, beyond preferences for specific sources acquired during pre- or post-training.
In the next section, we describe these issues and how we control for them through our data and methodology.

\section{Data and Methodology} %
\label{sec:data-methodology}

We opt for a dataset of synthetic knowledge conflicts (\S\ref{sec:setup-conflict-pairs}), with synthetic sources representing long-studied contrasts in credibility research (\S\ref{sec:setup-sources}).
Using this data, we evaluate $13$ open-weight models from four families (\S\ref{sec:setup-evaluation}).
The overall pipeline is shown in Figure \ref{fig:setup}.

\subsection{Plausible Knowledge Conflict Pairs}
\label{sec:setup-conflict-pairs}

We construct a dataset of equally plausible knowledge conflict pairs by perturbing attributes of fictional entities of seven types (\textsc{art}, \textsc{building}, \textsc{event}, \textsc{location}, \textsc{organization}, \textsc{person}, \textsc{product}), originally created in \texttt{NeoQA} to test out-of-domain QA 
rather than knowledge conflicts or source preferences 
\citep{glockner-etal-2025-neoqa}.
\texttt{NeoQA} entities are described with $38$ attributes such as \textit{date-of-birth} for \textsc{person} entities or \textit{headquarters} for \textsc{organization} entities.
This fictional data adheres to real world principles, shared units of measurements, and calendars, and is exhaustively validated with automatic and human checks.

Our conflict pairs consist of original \texttt{NeoQA} entities, and equally plausible counterfactual variants that differ in just one attribute value.
We generate four alternatives per entity attribute value.\footnote{We do not generate variations for \textit{name}, \textit{gender}, and \textit{spouse}, as the former is necessary for identifying the entity, and the latter two interact strongly with other attributes.} 
\textbf{Numerical attributes} (such as \textit{budget} or \textit{date-of-birth}) are automatically adjusted by up to $\pm 20\%$ or a fixed value depending on the attribute.
\textbf{Categorical attributes} with a small set of plausible values (such as \textit{marital status}) are sampled from a set of LLM-generated and manually-verified values.
Those with a large number of potential values such as \textit{profession} often depend on other entity attributes.
Here we generate alternatives for individual entities using \textsc{Qwen2.5-72B}.
Generation prompts and data creation details are provided in Appendix~\ref{app:conflictpairs}.

One author manually verified all created alternatives, correcting value formats and removing highly implausible instances (e.g., non-single \textit{marital status} for a child).
To maintain the dataset's synthetic nature, we remove proper noun values that have English Wikipedia articles.
With $373$ \texttt{NeoQA} entities, we create $1,903$ unique counterfactually-perturbed attribute values for a total of $7,440$ conflict pairs.

\subsection{Synthetic Sources}
\label{sec:setup-sources}

We create four types of fictional sources:

\paragraph{Newspaper.} We collect all U.S. newspaper names from Media Bias/Fact Check\footnote{\url{https://mediabiasfactcheck.com/}}, mask all location names using SpaCy \citep{spacy2020}, and extract the $150$ most frequent $2$-, $3$- and $4$-grams. After deduplication, $59$ newspaper templates such as \textit{"The \{LOC\} Herald"} remain. We fill these templates with fictional locations from \texttt{NeoQA} to create synthetic newspaper names.
    
\paragraph{Government.} Using \textsc{Qwen2.5-72B} we create templates for government agencies for each entity type (e.g., \textit{"Civil Registry of \{LOC\}"} for \textsc{person} entities). Again, we fill these with \texttt{NeoQA} locations.

\paragraph{Social media users.} We concatenate the @ symbol with random adjectives and nouns from WordNet\footnote{\url{https//wordnet.princeton.edu}} and four digits, mimicking Reddit's username suggestion algorithm (e.g., \textit{@GrantedMortal7505}). 
    
\paragraph{Person.} We sample the $200$ most frequent first and last names from the United States Census Bureau\footnote{\url{https//www.census.gov}} and Social Security Agency\footnote{\url{https://www.ssa.gov}} between 1945 and 2007.
We sample male and female names equally, and exclude combinations with an English Wikipedia page (e.g., \textit{Natalie Kennedy}) as before.

\subsection{Evaluation Method}
\label{sec:setup-evaluation}

\paragraph{Models.}
We evaluate $13$ instruction-tuned open-weight decoder-only models, covering a range of sizes and families, always presented in this order (from top to bottom) in figures:
\textsc{Qwen2.5} \textsc{7B}~\textcolor{orange}{$\blacksquare$}, \textsc{14B}~\textcolor{orange}{$\blacktriangle$}, \textsc{32B}~\textcolor{orange}{\ding{58}}, \textsc{72B}~\textcolor{orange}{$\bigstar$} \citep{qwen2025qwen25technicalreport},
\textsc{OLMo-2} 7B~\textcolor{darkgreen}{$\blacksquare$}, 13B~\textcolor{darkgreen}{$\blacktriangle$}, 32B~\textcolor{darkgreen}{\ding{58}} \citep{olmo20242olmo2furious}, 
\textsc{Llama-3.2} 3B~\textcolor{blue}{$\bullet$}, \textsc{Llama-3.1} \textsc{8B}~\textcolor{blue}{$\blacksquare$}
, \textsc{70B}~\textcolor{blue}{$\bigstar$}
\citep{grattafiori2024llama3herdmodels}, and
\textsc{Gemma-3} \textsc{4B}~\textcolor{red}{$\bullet$}
, \textsc{12B}~\textcolor{red}{$\blacktriangle$}
, \textsc{27B}~\textcolor{red}{\ding{58}}
\citep{gemmateam2025gemma3technicalreport}.

\paragraph{Forced-choice prompting.}
Each model input consists of an \textit{instruction}, a \textit{context}, a \textit{question}, and a set of \textit{answer options}.
The \textit{instruction} prompts the model to answer the subsequent multiple-choice question with an index token (e.g., \texttt{A} or \texttt{B}).
The \textit{context} contains a conflict pair from our dataset formatted as Markdown tables $T_A$ and $T_B$ to eliminate effects of text style
\cite{liu2025formatpriorquantifyinganalyzing}.
The pair is presented either without any source information (formalized as the tuple $C=(T_A, T_B)$), or with table A attributed to source instance $x$ of type $X$ and table B to a source instance $y$ of type $Y$ (formalized as $C'=(T_A^x, T_B^y)$).
For experiments comparing a source to no source, $x$ or $y$  in $C^\prime$ is the statement \textit{No source available}.
The \textit{question} then asks for an attribute value (e.g., \textit{nationality}) of an entity identified by name (e.g., \texttt{Sarah Kim}), using \textsc{Llama-3.1-70B}-generated and manually-verified templates.
Finally, the \textit{answer options} verbalize the conflicting attribute values copied from the tables with indices \texttt{A} and \texttt{B}.

To control for position bias \cite{zheng2023judging}, we use two versions of every prompt, also including $C_{rev}=(T_B, T_A)$ and $C'_{rev}=(T_B^y, T_A^x)$ (see Appendix~\ref{sec:position-bias}).
This results in a total dataset size of $2 \times 7,440$ data points.
We then obtain the deterministic probabilities of the answer tokens \texttt{A} and \texttt{B} to calculate the source preference metric;
we do not use generations, which have been shown to be ill-suited for investigating model preferences \citep{hu-levy-2023-prompting,subramonian2025agree}.
We extensively test the validity of our setup in Appendix~\ref{sec:setup-validation}, include example prompts in Appendix~\ref{sec:prompts-source-preferences}, and show that our results are stable under different prompts in Appendix~\ref{sec:prompt_stability}, following best practices \citep{sclar2024quantifying,mizrahi-etal-2024-state}.

\paragraph{Source preference metric.}
This metric quantifies the extent to which models' answers change when source information is introduced, isolating source preferences regardless of model-dependent preferences based on other parts of the prompts.
For each conflict pair, we first query a model $\theta$ for the probabilities of answer tokens \texttt{A} and \texttt{B} under the unattributed context $C$ and normalize them:
\begin{align*}
P_{\theta}(A|C)= \frac{P'_{\theta}(A|C)}{\sum_{x \in \{A,B\}} P'_{\theta}(x|C)}
\end{align*}

\noindent We then query the conflict under an attributed context $C'$ with sources $x$ and $y$ drawn from $X$ and $Y$, and compute $P_\theta(A|C')$ analogously.
We define a model's \textit{source preference} for a conflict pair as 
\begin{align*}
SP(\theta, T_A, x, T_B, y) = P_{\theta}(A|C')-P_{\theta}(A|C)
\end{align*}
A positive value indicates that $x$ increases the support for option $A$ more than $y$ supports $B$.
The aggregate measure $\widehat{SP}$ for source types $X, Y$ over a dataset $D$ of conflict pairs  averages SP over $D$, drawing instances of $X$ and $Y$ for every pair of conflicting tables: \\

\noindent \scalebox{0.75}{
$
\widehat{SP}(\theta; X, Y)
= \frac{1}{|D|}
\sum_{(T_A, T_B) \in D}
\left[
    SP(\theta, T_A, x, T_B, y),
    \begin{array}{l}
      x \in X\\
      y \in Y
    \end{array}
\right]
$
} \\

We visualize results with strip charts displaying $\widehat{SP}(\theta; X, Y)$, where $X$ is always the source on the right-hand side (RHS) of the chart.

\paragraph{Significance testing.}
We apply the nonparametric bootstrap test to our results with $n=10,000$, $\alpha = 0.05$, and Holm-Bonferroni correction.
In the rest of the paper, we only report results that are statistically significant for at least $10$ of $13$ models.

\section{LLM Source Preferences}
\label{sec:source-preferences}

We begin investigating LLM source preferences under knowledge conflicts by drawing from long-studied contrasts in credibility.
Specifically, we study the effects of source types of different presumed credibility (\S\ref{sec:inter-type-source-preferences}), as well as within-type features related to reputed credibility and sociodemographics (\S\ref{sec:intra-type-source-preferences}).
Then, we explore whether model behavior aligns with their source credibility judgments obtained through prompting \textit{without} knowledge conflict pairs (\S\ref{sec:introspection-source-preferences}), and whether preferences persist when models are allowed to abstain from answering (\S\ref{sec:abstention}).
We show example prompts in Appendix~\ref{sec:prompts-source-preferences}.

\subsection{Inter-Type Source Preference Behavior}
\label{sec:inter-type-source-preferences}

We first examine LLM preferences with four source types of varying presumed credibility: Governments, newspapers, social media users, and people.

\begin{figure}[t]
    \centering
    
    \includegraphics[width=\linewidth]{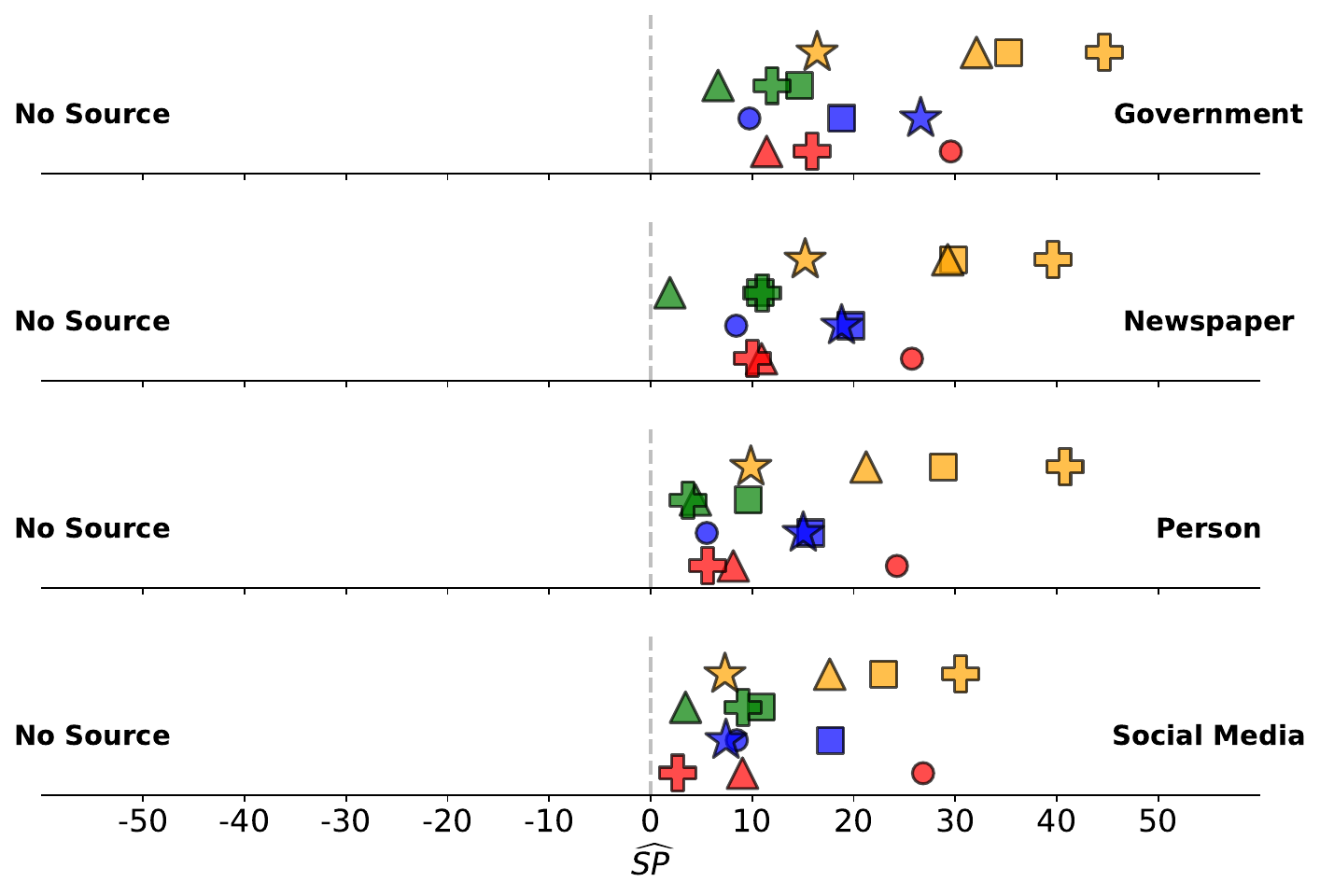}
    \caption{Source preferences when comparing attributed and non-attributed information: All models significantly prefer attributed information. Legend in \S\ref{sec:setup-evaluation}.}
    \label{fig:sc_attr}
\end{figure}

\begin{figure}[t]
    \centering
    \includegraphics[width=\linewidth]{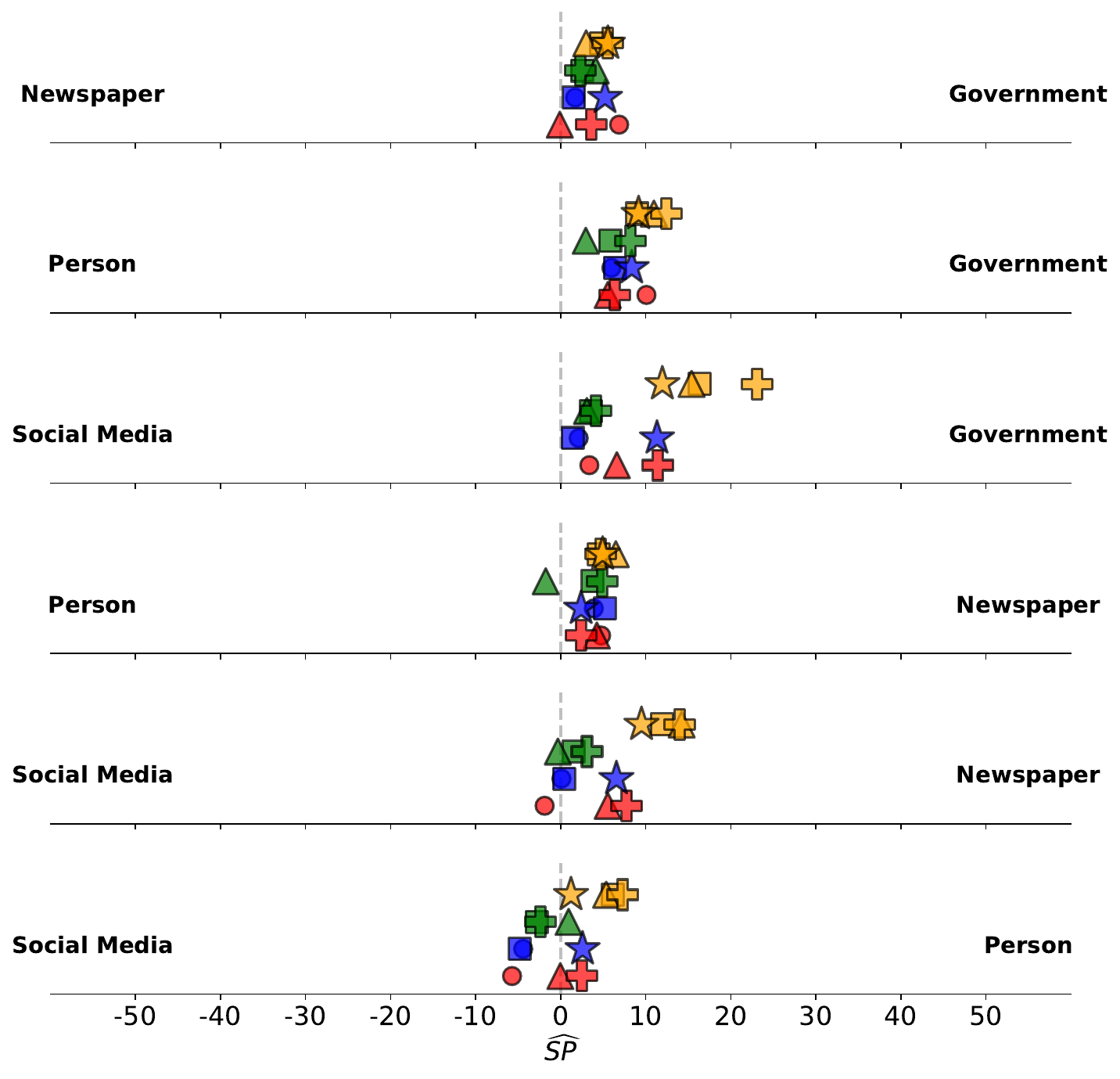}
    \caption{Model preferences between source types under knowledge conflicts: LLMs show strictly transitive preferences, aligning with an overall hierarchy of government $>$ newspaper $>$ individuals. Legend in \S\ref{sec:setup-evaluation}.}
    \label{fig:sc_st}
\end{figure}

Compared to \textit{No source available}, all models exhibit preferences for corroborated information across source types, as Figure~\ref{fig:sc_attr} shows.
When \textit{both} conflicting pieces of information are assigned sources of different types, all models show strictly transitive source preferences (see Figure~\ref{fig:sc_st}).
Inter-model source rankings of all four types are also highly consistent (average Kendall’s $W$ of 0.74);
11 of 13 models prefer both institutional sources over both individual sources.
Given the consistency of model rankings, we apply the single transferable vote algorithm with a Droop quota \citep{tideman-1995-stv} to induce a representative overall ranking across models, creating an \textbf{LLM credibility hierarchy where government $>$ newspaper $>$ person, social media}.
We also find that different methods of inducing this hierarchy are remarkably consistent (Appendix~\ref{sec:prompt_stability}).
Since we present all data instances in both orders, models must overcome position bias (Appendix \ref{sec:position-bias}) in order to display any source preference.
We find that position bias is negatively correlated with source preferences (-$0.4$ Spearman's $\rho$), indicating that our estimates are conservative and models' source preferences could be even stronger.

\subsection{Intra-Type Source Preference Behavior}
\label{sec:intra-type-source-preferences}

\begin{figure}[t]
    \centering
    \includegraphics[width=\linewidth]{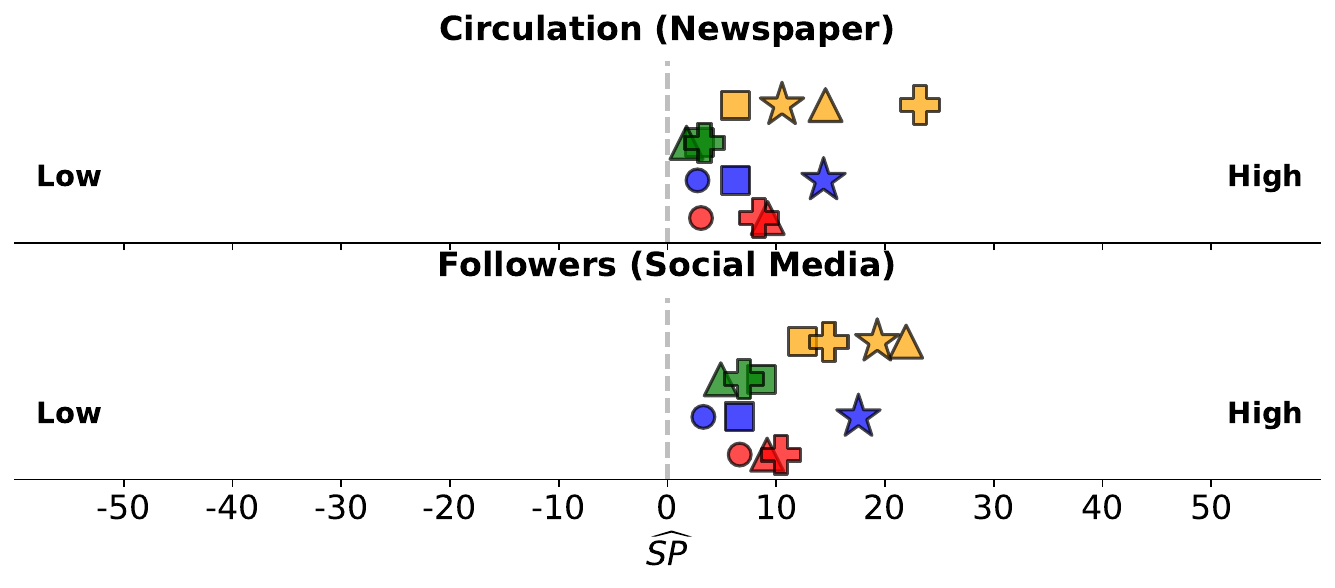}
    \caption{Preferences when conflicting information is attributed to sources of different reputed credibility (popularity): Models prefer popular sources. Legend in \S\ref{sec:setup-evaluation}.}
    \label{fig:sc_sa}
\end{figure}

Sources of the same type can still vary in  credibility.
Thus, we study source  properties  related to reputed credibility \cite{fogg-tseng-1999-elements} and sociodemographics, and their impact on how LLMs resolve knowledge conflicts.
Details on data construction and more fine-grained  results are in Appendix~\ref{sec:intra_details}.
Motivated by the impact of newspaper circulation \cite{meyer2004influence} and social media reach  \cite{waddell2018does, morris2012tweeting} on credibility, we investigate \textbf{source popularity} via circulation numbers for newspaper  and follower counts for social media sources.
As Figure~\ref{fig:sc_sa} shows, all models tend to resolve conflicts based on higher source popularity.

\begin{figure}[t]
    \centering
    \includegraphics[width=\linewidth]{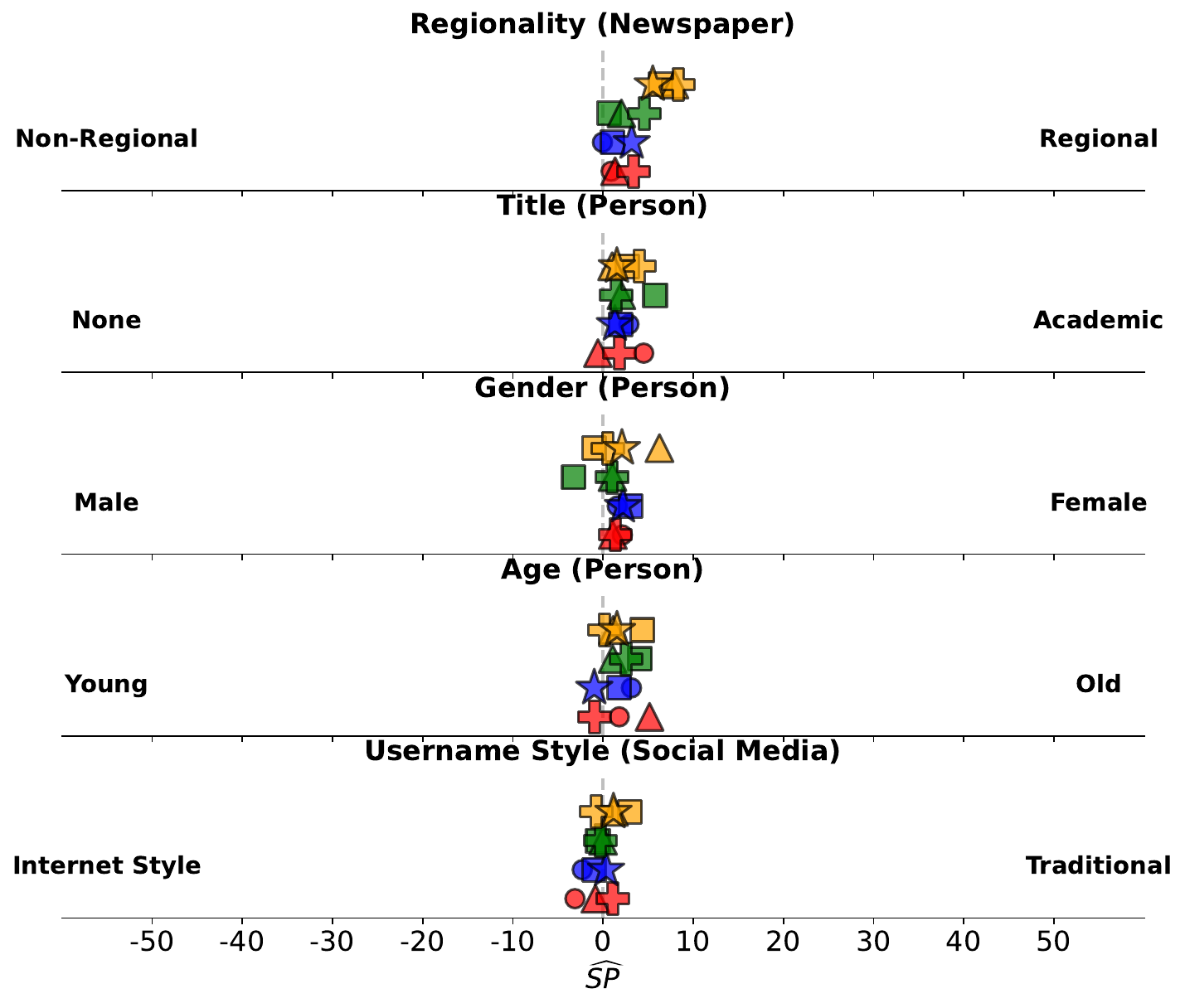}
    \caption{Source preferences when conflicting information is attributed to sources with different sociodemographic characteristics: Most models slightly prefer regional sources, academic titles, women, and older people, while username style is mixed. Legend in \S\ref{sec:setup-evaluation}.}
    \label{fig:social}
\end{figure}

Next, we examine how sociodemographic factors affect source preferences, as they are well-known to affect NLP systems in other contexts \citep{gallegos-etal-2024-bias}.
Motivated by U.S. adults' higher trust in local over national news \cite{pewresearch2025how,fioroni2022local}, we consider newspapers' \textbf{regional proximity to the entity}, by comparing fictional sources with the same location as the entity, to sources with a different location.
Next, we augment people sources with \textbf{academic titles, gender and age};
among humans, academic titles confer higher credibility \citep{yan-2023-trust,NowakKramer2025}, as does being male \citep{NowakKramer2025,Weibel08082008}, while results on age are mixed \citep{mueller-johnson-et-al-2007-ageism,Weibel08082008}.
As for social media users, we investigate whether LLMs trust \textbf{traditional usernames} (composed of first and last names connected with underscores or camel case, e.g., \textit{@Joshua\_Reyes}) over our previously constructed internet usernames, as humans do \citep{morris2012tweeting}.
Surprisingly, Figure~\ref{fig:social} shows that sociodemographic factors have only a weak  effect on behavioral LLM source preferences.

\subsection{Prompted Preferences vs. Behavior}
\label{sec:introspection-source-preferences}

We have shown that models display clear source preferences under knowledge conflicts.
Prior research has studied the ability of LLMs to directly assess source credibility via context-independent prompting \citep{yang2025accuracy, prama-islam-2025-evaluating}.
Given that differences between prompting and behavior have also been studied in stereotypes, grammaticality, and value alignment \citep{goldfarb-tarrant-etal-2021-intrinsic,hu-levy-2023-prompting,shen-etal-2025-mind}, we test whether our behavioral source preferences align with prompted preferences.
We elicit credibility judgments on  source pairs from different types with prompts like: \textit{"Who would you rate higher in terms of credibility? (A) Melvidra News Press (B) Matthew Chen"}.
For every prior source matchup (except regionality, which requires context),
we construct $2,000$ inputs, consisting of $1,000$ source pairs presented in both orders, each paired with a question sampled from $20$ templates (see Appendix~\ref{sec:prompts-source-preferences} for prompts and questions).
We report probability deviation from 50\% for answer tokens representing each source.

Figure~\ref{fig:expl} exemplifies broad patterns with prompted preferences (see Appendix~\ref{sec:fullstated} for all $15$ source contrasts $\times$ $13$ models = $195$ cases).
Prompting mostly  ($139$ out of $195$ cases) elicits significantly stronger preferences in the same direction as model behavior.
However, models flip in $38$ cases from significant preferences in one direction to the opposite.
These tend to be previous outliers, e.g., \textsc{Gemma-3-27B} and \textsc{Llama-3.1-70B} were the only models to prefer young  over old people in their behavior, but they flip when prompted.
These flips lead to more consistent preferences:
Inter-model agreement (Kendall's $W$) goes up from $0.59$ to $0.77$ with prompting.
While prompting produces more dramatic contrasts, behavioral evaluation remains more consistent with model use.

\begin{figure}[t]
    \centering
    \includegraphics[width=\linewidth]{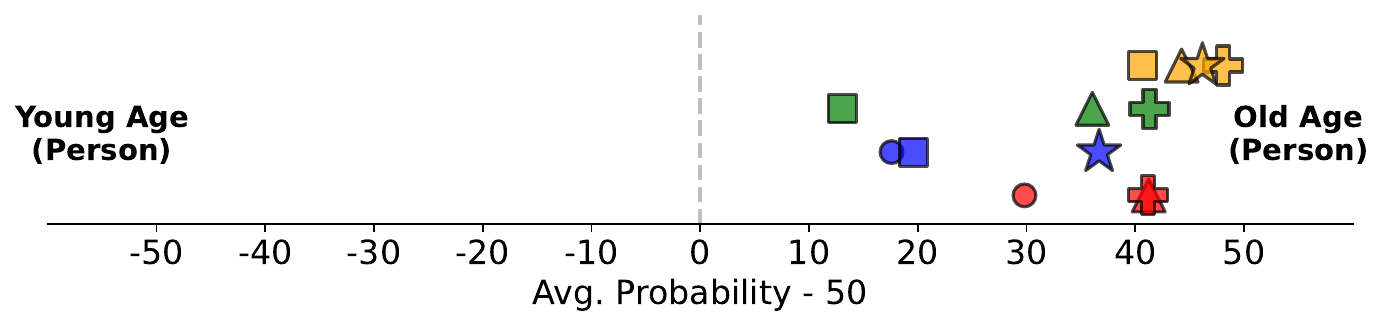}
    \caption{Probability deviation from 50\% of  RHS answer  when models are directly prompted to choose the more credible source without context. Legend in \S\ref{sec:setup-evaluation}.}
    \label{fig:expl}
\end{figure}

\subsection{Abstention Under Conflicting Contexts}
\label{sec:abstention}

To evaluate whether source preferences only emerge through forced-choice prompting, where models must select an answer, we repeat our experiments from \S\ref{sec:inter-type-source-preferences} with a third answer option for models to abstain or express uncertainty (\textit{"Can't be answered"}).
We use numbers instead of letters as answer tokens (see Appendix~\ref{sec:different_tokens}) to decouple table indices from answer tokens, and prompt with all $6$ possible table and answer orders. 
We measure the probability assigned to the abstain option in $C$ and $C'$ inputs alongside $\widehat{SP}$ of the conflicting source types (which now no longer sum up to 0). 
In line with prior findings \citep{madhusudhan-etal-2025-llms,tam-etal-2025-none}, Table~\ref{tab:neutral_option} shows that most models do not abstain from answering in the majority of cases, despite having too little information to resolve the presented conflict.
Counterintuitively, most models show just marginal differences when further information in the form of attributed sources is presented.
However, even in this three-option setting, source preferences induced from the higher $\widehat{SP}$ values are still fully transitive for all models, with $11$ out of $13$ models producing the same hierarchy as in our forced-choice setup.
This demonstrates that source preferences affect model resolution even in scenarios with the option to abstain.

\begin{table}[]
\scalebox{0.93}{
\begin{tabular}{lrr}
\hline
\textbf{Model} & \multicolumn{1}{c}{\textbf{\begin{tabular}[c]{@{}c@{}}Avg. Prob. of \\ Neutr. in $C$\end{tabular}}} & \multicolumn{1}{c}{\textbf{\begin{tabular}[c]{@{}c@{}}Avg. Prob of \\  Neutr. in $C'$\end{tabular}}} \\ \hline
\textsc{Qwen2.5-7B}     & 2.0 \%                                                                                          & 2.8 \%                                                                                                 \\
\textsc{Qwen2.5 14B}    & 46.2\%                                                                                          & \textbf{55.9\%}                                                                                        \\
\textsc{Qwen2.5 32B}    & \textbf{59.4\%}                                                                                 & 45.8\%                                                                                                 \\
\textsc{Qwen2.5 72B}    & 29.8\%                                                                                          & 40.2\%                                                                                                 \\ \hline
\textsc{OLMo-2 7B}      & 0.2\%                                                                                           & 0.2\%                                                                                                  \\
\textsc{OLMo-2 13B}     & 1.0\%                                                                                           & 1.3\%                                                                                                  \\
\textsc{OLMo-2 32B}     & 0.2\%                                                                                           & 0.2\%                                                                                                  \\ \hline  
\textsc{Llama-3.2 3B}   & 2.9\%                                                                                           & 3.8\%                                                                                                  \\
\textsc{Llama-3.1 8B}   & 1.5\%                                                                                           & 1.8\%                                                                                                  \\
\textsc{Llama-3.1 70B}  & 43.9\%                                                                                          & 41.5\%                                                                                                 \\ \hline    
  
\textsc{gemma-3 4B}     & 1.3\%                                                                                           & 1.6\%                                                                                                  \\
\textsc{gemma-3 12B}    & 9.1\%                                                                                           & 12.1\%                                                                                                 \\
\textsc{gemma-3 27B}    & 20.0\%                                                                                          & 9.5\%                                                                                              \\ \hline    

\end{tabular}
}
\caption{Average probability assigned to a neutral option \textit{"Can't be answered"}. Most models only abstain on the minority of inputs, whether source information is present ($C'$) or not  ($C$). Exceptions are bolded.}
\label{tab:neutral_option}
\end{table}

\section{Credibility vs. Majority vs. Repetition}
\label{sec:repetition-bias}

So far, we have studied LLM source preferences  in isolation, but they might also interact with other preferences;
in other work on knowledge conflicts, \citet{xie2024adaptive} and \citet{jin2024tug} have shown that LLMs tend to follow the majority, similar to the bandwagon effect in humans \cite{leibenstein1950bandwagon}.
In contrast to prior work, we disentangle majority and repetition bias, and examine their interaction with source preferences.
We operationalize this by comparing \textit{government} minority and \textit{social media} majority sources in three ways, as they are the correspondingly most and least preferred sources. We give example prompts in Appendix~\ref{sec:prompt_ex3}:

\paragraph{2-Table Majority:} Three tables are shown separately, of which two identical ones  are attributed to two \textit{different} social media sources $x1, x2 \ (x1 \neq x2)$, and the conflicting one is attributed to a government source $y$, i.e., $C' = \left( T_A^{x1},\; T_A^{x2},\; T_B^{y} \right)$.

\paragraph{1-Table Majority:} The two agreeing social media sources are merged in the header of a single table, so no table is repeated: $C' = \left( T_A^{x1,x2},\; T_B^{y} \right)$.

\paragraph{Repetition:} Three tables are shown separately, of which two identical ones are attributed to the \textit{same} social media source $x1$ and the conflicting one is attributed to a government source $y$. More formally, $C' = \left( T_A^{x1},\; T_A^{x1},\; T_B^{y} \right)$. \\

\begin{figure}[t]
    \centering
    \includegraphics[width=\linewidth]{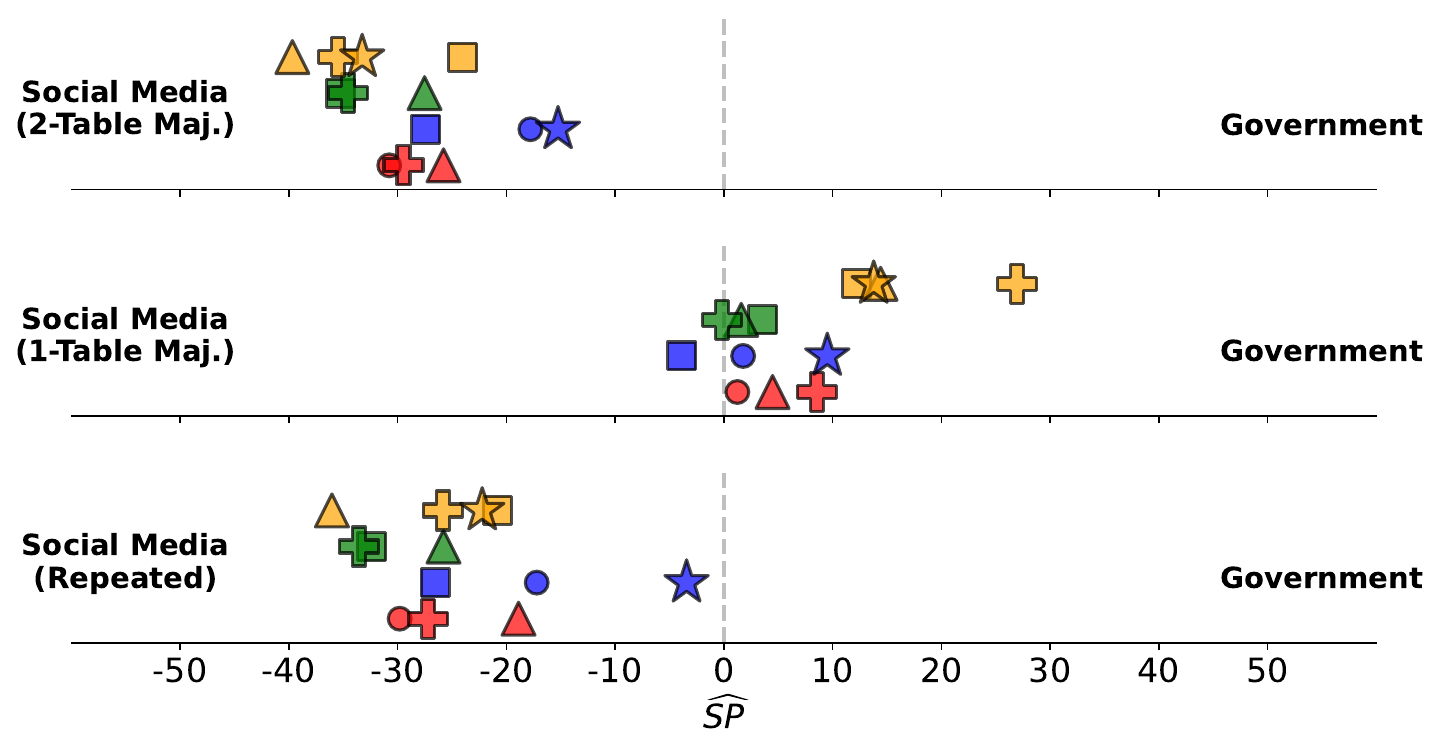}
    \caption{Preferences contrasting a majority/repetition of previously low-credibility sources with a previously high-credibility authority in three settings. Repeated information (whether attributed to a single source or two different ones) flips prior rankings. Legend in \S\ref{sec:setup-evaluation}.}
    \label{fig:major2}
\end{figure}

We evaluate all combinations of context and answer orders, and report the $\widehat{SP}$ gap, i.e.,
the absolute difference between $\widehat{SP}$ with repetition or majority, and $\widehat{SP}$ without it, in an otherwise equal setting.
Figure~\ref{fig:major2} shows that with a 2-Table Majority, all models prefer the previously low-credibility social media sources, with an average $\widehat{SP}$ gap of $33.90$.
However, when the same majority is presented in the 1-Table setting, models stick with their original government preference (average $\widehat{SP}$ gap of only $6.17$).
The Repetition setting lets us disentangle whether we find  a majority bias or simply a preference for repeated tokens.
Indeed, all models prefer repeated information (average $\widehat{SP}$ gap of $30.04$), even though no new source is provided and thus no true majority presented.
This reveals a clear vulnerability of LLMs to repeated disinformation \citep{NewsGuard_2024} and can be seen as a correlate of the illusory truth effect in humans; a single repetition of true or false information leads to humans evaluating it as more accurate \cite{hasher1977frequency, fazio2015knowledge, pennycook2018prior}, even overruling source credibility \cite{begg1992dissociation}.

\begin{figure}[t]
    \centering
    \includegraphics[width=\linewidth]{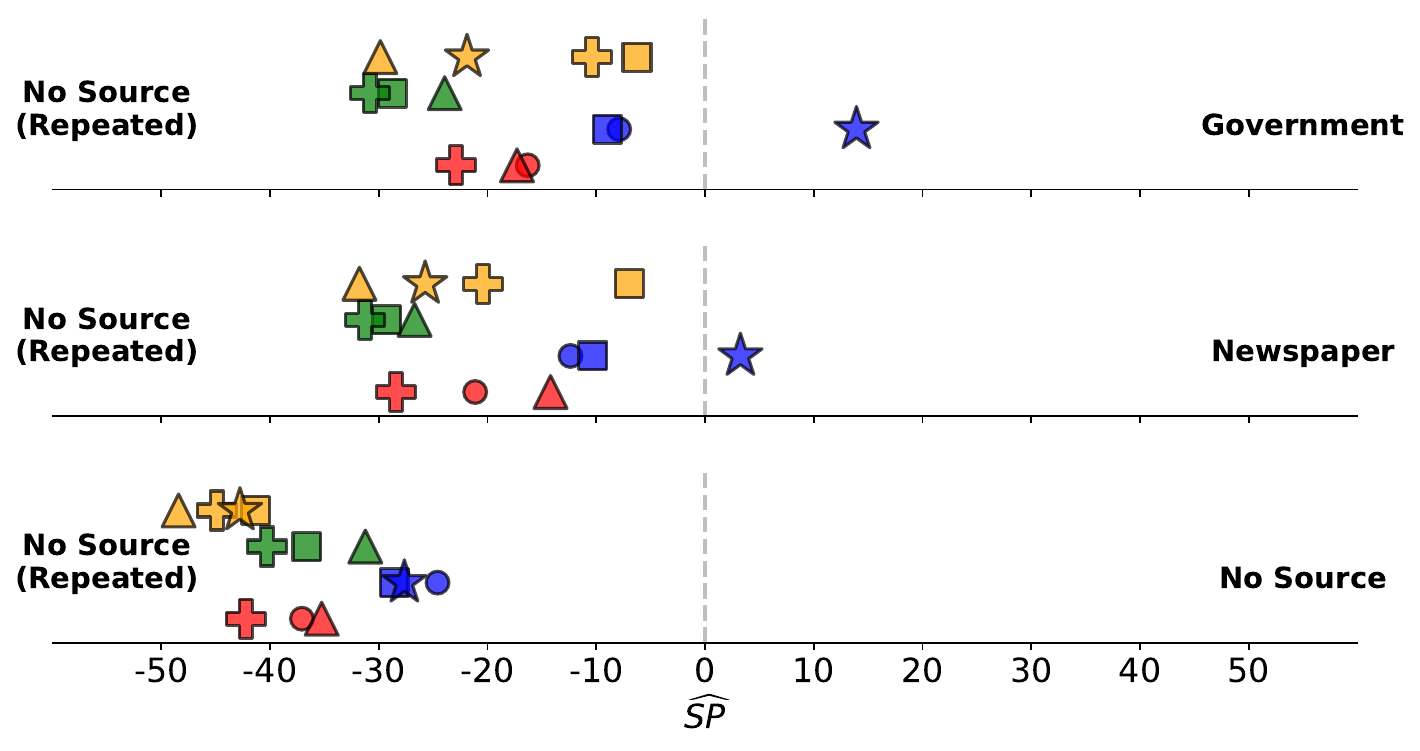}
    \caption{LLMs mostly prefer \textit{repeated} unattributed information, flipping prior preferences for attributed information. Legend in \S\ref{sec:setup-evaluation}.}
    \label{fig:major1}
\end{figure}

Figure~\ref{fig:major1} shows that repetition bias persists even when we repeat \textit{unattributed} information, flipping the original preferences in Figure~\ref{fig:sc_attr}.
When no source is provided in either case (last row of Figure~\ref{fig:major1}), repetition bias is even stronger, indicating that source credibility still plays a role in this setting, but takes a backseat compared to repetition.

\begin{figure}[t]
    \centering
    \includegraphics[width=\linewidth]{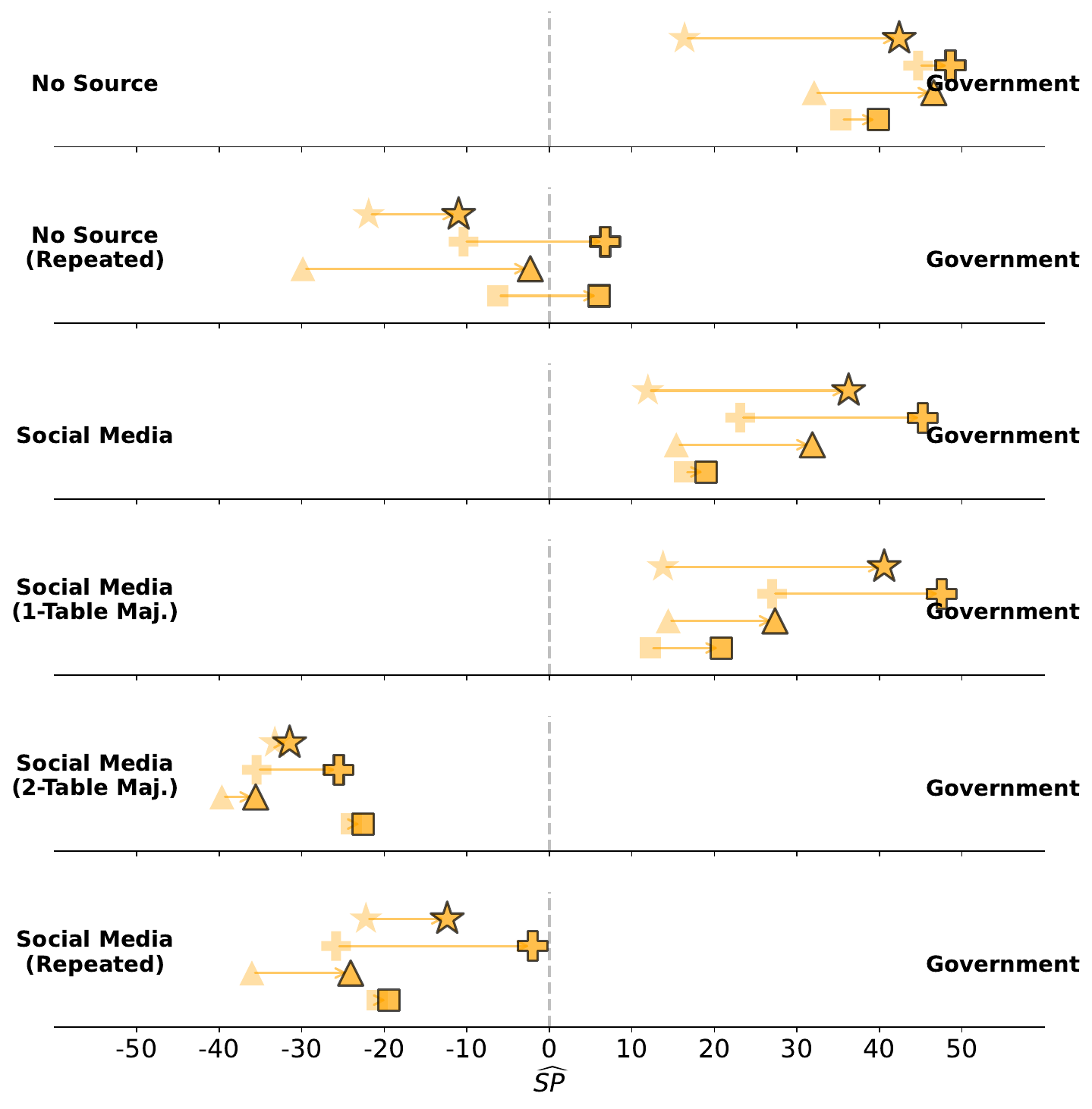}
    \caption{Source preferences  when \textsc{Qwen} models are instructed to consider source credibility (darker), compared to original prompts (lighter). 
    Prompting weakens repetition bias but not enough to ensure consistency with the original source hierarchy. Legend in \S\ref{sec:setup-evaluation}.}
    \label{fig:promptbased_qwen}
\end{figure}

\paragraph{Repetition prompting.}
We investigate whether repetition bias can be reduced by  prompting models to attend to source credibility and repetition.
We add the following to the \textit{instruction}: \textit{"When selecting an answer, identify which sources support each option and assess the credibility of those sources before deciding. Simple repetition of information should not influence your choice."}
Figure~\ref{fig:promptbased_qwen} shows results for \textsc{Qwen} models (see  Appendix~\ref{sec:full_prompt_based_repetition} for others).
The rightward shifts show that this does strengthen original source preferences, with a greater effect at mitigating repetition bias compared to a 2-Table true majority.
However, in most cases, prompting is insufficient to ensure consistency with the original source hierarchy.

Since repetition-aware prompting might not be well-suited to simulate real LLM use, we include another scenario where users only ask models to consider the sources.
Results of this credibility-aware prompting are shown in Appendix~\ref{sec:full_prompt_based_cred}.

\section{Mitigating Repetition Bias}
\label{sec:mitigation}

While the question of which source preferences LLMs \textit{should} have is complex (see our \nameref{sec:ethics}), repetition bias is clearly dangerous, as it renders models vulnerable to adversarial attacks.
Ideally, models should remain consistent with their original source preferences even with repetition, which repetition-aware prompting does not accomplish.
Therefore, we propose a teacher-student knowledge distillation paradigm to minimize
differences in model preferences between inputs with and without repeated information.
Additional details for replication are in Appendix \ref{sec:mitigation-details}.

\paragraph{Training and test data.}
We train on $1500$ conflict pairs of $12$ seed entities and all their counterfactual variations, combined with newspaper and person sources.
Each example consists of an aligned pair of prompts $(C'_U, C'_R)$:
$C'_U$ where $C'_U=(T_A^x, T^y_B)$ and $C'_R$, where we randomly repeat one of the tables.
We evaluate the final model's $\widehat{SP}$ on a held-out test set of all conflict pairs from remaining $361$ entities combined with the previously unseen government and social media sources.

\paragraph{Training objective.}
Let $f_t$ be the frozen base model (the teacher), and $f_s$ be the same model with LoRA parameters (the student).
For a pair $(C'_U,C'_R)$, we obtain both models' normalized token probabilities \textit{A} and \textit{B}.
We optimize $f_s$ with a weighted loss ($\lambda=0.2$) composed of two Kullback–Leibler divergences as in \citet{qiang-etal-2024-prompt}; the first term constrains $f_s$ to mimic the base model in settings with no repetition, the second penalizes deviations with repeated information:
\begin{align*}
    \mathcal{L}
=
\lambda &D_{KL}\!\left(f_t(C'_U) \,\|\, f_s(C'_U)\right) \\
+ (1-\lambda)&D_{KL}\!\left(f_t(C'_U) \,\|\, f_s(C'_R)\right)
\end{align*}

\paragraph{Model and optimization setup.}
We fine-tune \textsc{Gemma-3-4B}, one of our smallest models, using LoRA \citep{hu2022lora} with conventional parameters.
After this, we train for $200$ steps on just the first loss term to further retain the original source preferences.
We report averages of 7 seeds with randomly generated held-out training entities each.

\paragraph{Results}

\begin{figure}[t]
    \centering
    \includegraphics[width=\linewidth]{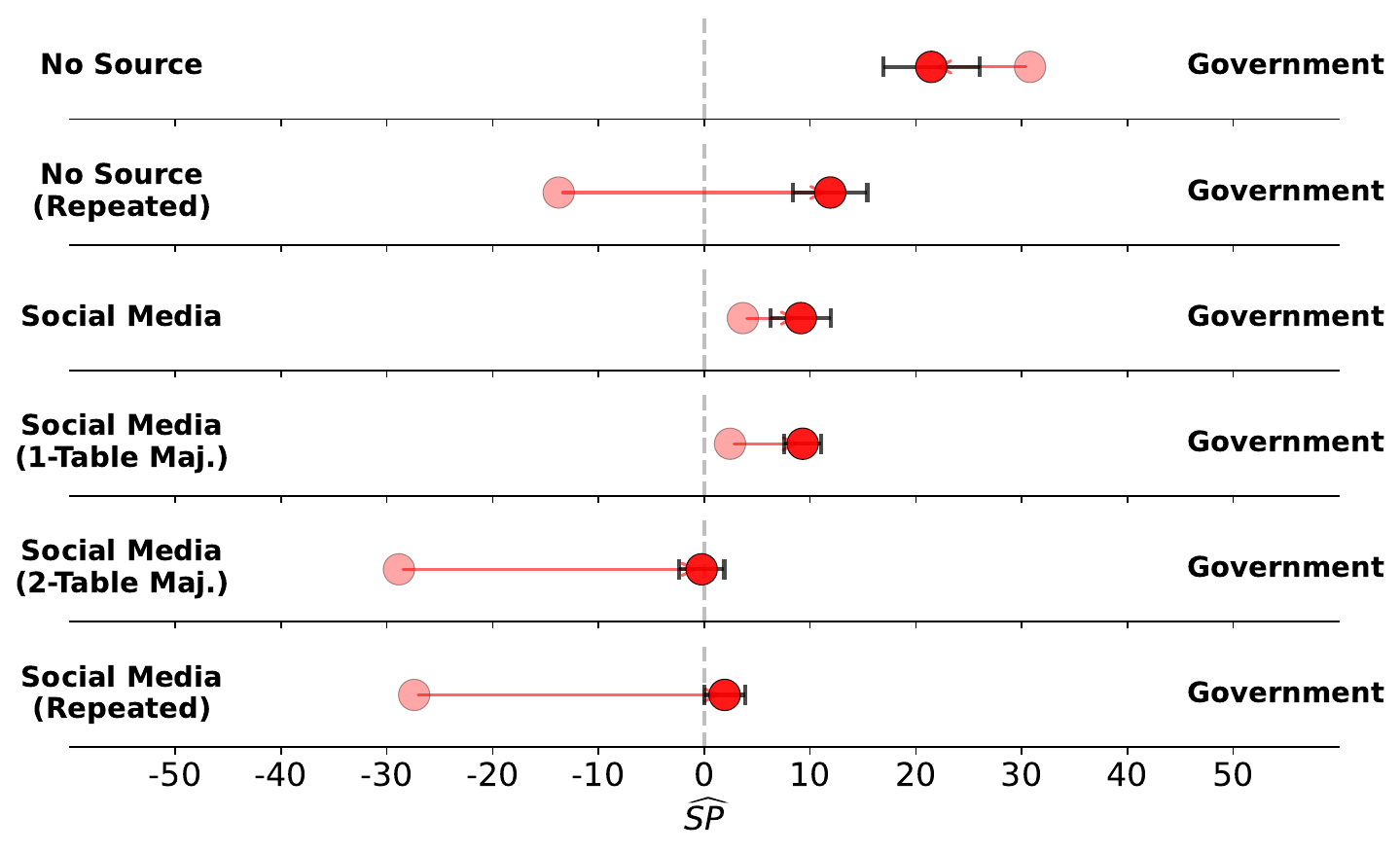}
    \caption{\textsc{Gemma3-4B}s original teacher model (lighter) compared to the average fine-tuned and credibility-prompted model (darker) with standard deviation across $7$ different training runs. This setup reduces repetition bias and maintains original preferences.}
    \label{fig:mitft}
\end{figure}

As Figure~\ref{fig:mitft} shows, the combination of fine-tuning and a credibility-aware prompt (Appendix \ref{sec:full_prompt_based_cred}) successfully mitigates repetition bias, while mostly retaining the  (teacher model's) no-repetition source preferences, generalizing to source types not seen during training.
The $\widehat{SP}$-gap between government and \textit{no source}
reduces by $79.2\%$ from $45.9$ to $9.55$, while retaining $72.5$\% of the original preference from $29.6$ to $21.5$.
For government and social media conflicts, the reduction in repetition bias is $78.3\%$, from $33.2$ to $7.2$.
Here, the preference for government increases slightly from $3.4$ to $9.1$.
This also leads to more similar preferences between 1-Table and 2-Table majorities.
Thus, models can in fact be trained to behave consistently under repetition.

\section{Related Works}
\label{sec:related-work}

\paragraph{Knowledge conflicts.}
Work on knowledge conflicts with LLMs (surveyed in \citet{xu2024knowledge}) began with conflicts between parametric and contextual knowledge \citep{longpre-etal-2021-entity,xie2024adaptive, jin2024tug} in QA and RAG.
Subsequent work has explored conflicts within parametric knowledge \citep{su2024textttconflictbank,marjanovic-etal-2024-dynamicqa}, as well as within contexts, which is our setting \citep{li-etal-2024-contradoc,liu-etal-2025-open}.
Entity swaps~\citep{gautam-etal-2023-lightweight} are commonly used to create synthetic conflicts in this literature, as we do.
Closest to our work, \citet{kurfali-2025-conflicting} study the effects of repetition and position of conflicts in long-context retrieval, and
\citet{shaier2024adaptive} teach LLMs to cite their sources for all possible answers to address conflicts in open-ended QA.
However, to the best of our knowledge, no work prior to ours considers source credibility.

\paragraph{Credibility.}
Within NLP, credibility has been used in the context of assessing information quality to detect argument quality \citep{walker-etal-2018-evidence}, rumours \citep{li-etal-2019-rumor-detection}, fake news \citep{yuan-etal-2020-early}, and low-quality science \citep{augenstein-2021-determining}.
Within QA and RAG, there is less grounding in extra-disciplinary credibility research.
\citet{wan-etal-2024-evidence} studies what (but not whose) evidence models find convincing, examining relevance and style.
\citet{shen2025transparentize} also consider aspects of message credibility such as style, conciseness and logical consistency.
As in our work, \citet{hong-etal-2024-gullible} use prompting and fine-tuning to mitigate RAG sensitivity to noisy information.
\citet{dai-etal-2025-media} show that LLM citation preferences for news articles are strongly influenced by political bias in source selection.
Similarly, \citet{khan2026in} find that LLM agents exhibit systematic latent intra-type source preferences across news, academic, and e-commerce domains.
Finally, \citet{pan-etal-2024-contexts} train models to incorporate source credibility information during generation, in contrast to our evaluation of models' inherent judgments.

\paragraph{Biases.}
NLP technologies have been shown to display similar cognitive biases to humans \citep{malberg-etal-2025-comprehensive}, of which the bandwagon effect \cite{xie2024adaptive, jin2024tug}, illusory truth effect \cite{griffin2023large} and authority bias \citep[studied in the context of LLM-as-a-judge;][]{ye2025justice,wang2025assessing, chen2024humans} are relevant to our work.
None of the above work studies these biases in the context of source credibility in inter-context knowledge conflicts.
\citet{malaviya-etal-2022-cascading} show how such biases may emerge from cognitive shortcuts by human annotators, and
\citet{mina-etal-2025-cognitive} analyze the interplay of multiple cognitive biases, similar to
our study of the interaction between source credibility and bandwagon or illusory truth effects.
The combination of repetition bias and stereotypical (gender) biases also influences LLM behavior under conflicts in language modeling \citep{gautam-et-al-2024-robust}.
Finally, format biases, which we do not study, affect LLM behavior under conflicts in RAG \citep{liu2025formatpriorquantifyinganalyzing}.

\section{Conclusion}
Through extensive experiments in a synthetic setting designed to isolate LLM source preferences, we find that characteristics of the source affect how models resolve inter-context knowledge conflicts.
Models show clear hierarchical preferences for sources with higher presumed and reputed credibility, which even persist when models are presented the option to abstain.
Model behavior differs from human-like credibility judgments under sociodemographic variation, and directly-prompted preferences vary significantly from actual behavior.
We disentangle repetition and majority biases and show that repeated information can flip source preferences, and  prompting cannot sufficiently mitigate this.
Finally, we propose a novel fine-tuning method which, when combined with credibility prompting, teaches models repetition invariance and preserves original source preferences.

\section*{Limitations}

\paragraph{Synthetic setting.}
We focus on entirely synthetic scenarios in order to isolate source effects in inter-context conflicts, which are hard to measure with confounds from parametric knowledge.
Although we consider our trust hierarchy in Section~\ref{sec:source-preferences} to be representative of real-world scenarios as well, there are particular examples where this may not hold, modulated by the style and topic of the message, as well as the expertise of the source.
Even in one of the oldest studies on credibility \citep{hovland-weiss-1951-influence}, an individual (\textit{J. Robert Oppenheimer}, a famous American theoretical physicist) is shown to be more credible to U.S. participants than a newspaper (\textit{Pravda}, a Russian broadsheet newspaper) on the subject of atomic submarines.
Similarly, we propose advanced solutions for the problem of repetition bias in Section \ref{sec:mitigation}, despite our simple setting where deduplicating the knowledge base would also work.
However, in realistic RAG systems, it would neither be as trivial to deduplicate information as it is in our synthetic setting, nor would it be appropriate to do so in a source-agnostic way.
Our mitigation strategy is source-aware, but we do not know if it generalizes to different knowledge conflicts, contexts and tasks in more realistic settings.

\paragraph{Evaluation strategy.}
We experiment exclusively with a forced-choice question answering setup, i.e., no step-by-step reasoning, and no generative answers.
We chose this setup to simplify evaluation while remaining true to common RAG setups \citep{lewis2020retrieval}, but note that alternate evaluation strategies could produce different results \citep{hu-levy-2023-prompting,tam2024let,chen2024two,subramonian2025agree}.

\paragraph{Content domain.}
Our tasks are designed to focus on conflicts in factual content, with no sentiment-based or preferential questions that introduce ambiguity.
Furthermore, our conflicts present equally plausible alternatives, unlike real-world data, where there may be conflicts between a priori more and less plausible alternatives (e.g., information about the moon landing vs. conspiracies about the moon landing), and where certain sources may have more epistemic authority than others (e.g., \textit{NASA} about the moon landing).
We choose to present the data in tabular form to reduce the effects of message style, which is  known to affect credibility judgments in humans as well \citep[surface credibility;][]{fogg-tseng-1999-elements}.
We leave it to future work to investigate how these other aspects of the content interact with source credibility.

\paragraph{Language and culture.}
We experiment only with English language data and prompting, and we use U.S. preferences in some aspects of our experimental setup (e.g., our choice of names, our newspaper templates).
Although some aspects of source credibility may be similar across cultures \citep{yoon-et-al-1998-cross-cultural}, this is not always true \citep{morimoto-la-ferle-2008-examining}.
Therefore, it is likely that other languages may evoke different credibility behavior and a potentially different trust hierarchy than the one we present in Section \ref{sec:source-preferences}.

\section*{Ethics Statement}
\label{sec:ethics}

In this paper we take a descriptive rather than a prescriptive view of source credibility, as there exists no perfect, context-free source hierarchy that we should align models to.
We take the normative position that institutional trust is generally good \citep{estadieu-2025-institutional-trust}, but note that institutions can be captured and lobbied \citep{bo-2006-regulatory-capture}.
Additionally, institutional power dynamics typically replicate societal power dynamics along lines of race, gender, and so on;
thus, these are areas where individual marginalized voices can be more credible than the institutional view, where they may get drowned out \citep{crenshaw-1991-mapping}.
Finally, we emphasize that we do not and do not wish to anthropomorphize large language models despite studying LLM credibility judgments \citep{PROUDFOOT2011950}.
We take the position that human credibility preferences are reflected in training data and thus implicitly learned by language models, but this does not make them entities that have preferences themselves or that can ``introspect'' on their beliefs.

\section*{Acknowledgments}
The authors acknowledge support by the German state of Baden-Württemberg through bwHPC and the German Research Foundation (DFG) through grant INST 35/1597-1 FUGG.

\bibliography{custom}
\clearpage

\appendix

\begin{table*}[th!]
\begin{tabularx}{\linewidth}{lllXl}
\hline
\textbf{Entity Type} & \textbf{Attribute} & \textbf{\texttt{NeoQA} Value} & \textbf{Alternatives}                                                                                        & \textbf{Method} \\ \hline
\textsc{Person}               & \textit{Eye color}          & \texttt{Blue}                 & \texttt{Hazel}, \texttt{Green}, \texttt{Black}, \texttt{Brown}                                                                                   & Sampling        \\ \hline
\textsc{Person}               & \textit{Marital status}     & \texttt{Married}              & \texttt{Single}, \texttt{Widowed}, \texttt{Divorced}, \texttt{Engaged}                                                                           & Sampling        \\ \hline
\textsc{Event}                & \textit{Date}              & \texttt{2023-11-10}           & \texttt{2022-12-31}, \texttt{2024-04-26}, \texttt{2024-11-02}, \texttt{2024-09-10}                    & Rescaling       \\ \hline
\textsc{Building}             & \textit{Capacity}           & \texttt{250}                 & \texttt{280, 270, 290, 230}                                                                                        & Rescaling       \\ \hline
\textsc{Location}             & \textit{Country}            & \texttt{Asvelia}              & \texttt{Breloria}, \texttt{Eldoria}, \texttt{Nvestale}, \texttt{Bremorin}, \texttt{Thysvelia}                  & Generation      \\ \hline
\textsc{Organization}        & \textit{Industry}           & \texttt{Public Oversight}          & \texttt{Ethical Technology Regulation}, \texttt{Digital Privacy Advocacy}, \texttt{Innovation Compliance Monitoring}, \texttt{Autonomous Systems Governance} & Generation      \\ \hline
\end{tabularx}
\caption{Examples of counterfactual alternatives to \texttt{NeoQA} attribute values}
\label{tab:altexs}
\end{table*}

\section{Creation of Conflict Pairs}
\label{app:conflictpairs}

In Table~\ref{tab:altexs}, we show examples of four counterfactually-created alternative values for different entity types and attributes.
In the following subsections, we describe three different methods of creating counterfactual alternatives in more detail:
\begin{enumerate}
    \item \textbf{Rescaling} for numerical attributes (Appendix \ref{sec:rescaling})
    \item \textbf{Sampling} for categorical attributes with a small number of possible values (Appendix \ref{sec:sampling})
    \item \textbf{Generation} for categorical attributes with a large number of possible values (Appendix \ref{sec:generation})
\end{enumerate}

\subsection{Rescaling}
\label{sec:rescaling}

We automatically adjust the values of numerical attributes that are not dates (such as \textit{budget}) by up to $\pm 20\%$.
Numbers with five digits or more are rounded to the third most significant decimal place to preserve a consistent level of precision.
We scale dates that only consist of years by up to $\pm 30$ years, while staying within the range $1850-2025$, close to the \texttt{NeoQA} values.
We rescale exact dates by up to $\pm 365$ days.

\subsection{Sampling}
\label{sec:sampling}

For attributes with a small set of plausible values, we prompt OpenAI's ChatGPT via the web interface, using \textsc{GPT-4.1} \citep{openai2024gpt4technicalreport} to create sets of alternative values, which we manually filter.
The prompt used for creating the sets of values is shown in Figure \ref{p:sample}.
When creating perturbed alternatives for an attribute, we sample one value from the corresponding set.

\paragraph{Variations for the \textit{material} attribute of \textsc{building} entities} are \texttt{stone and timber},
\texttt{brick and wood},
\texttt{steel and concrete},
\texttt{glass and aluminum},
\texttt{bamboo and steel},
\texttt{limestone and glass},
\texttt{sandstone and oak},
\texttt{ceramic and metal},
\texttt{slate and pine},
\texttt{brick and concrete},
\texttt{glass and steel},
\texttt{wood and concrete},
\texttt{stone and glass},
\texttt{timber and concrete},
\texttt{brick and stone},
\texttt{wood and aluminum},
\texttt{concrete and aluminum},
\texttt{glass and timber},
\texttt{stone and steel},
\texttt{brick and steel},
\texttt{concrete and glass},
\texttt{timber and glass},
\texttt{brick and timber},
\texttt{stone and concrete},
\texttt{wood and steel},
\texttt{glass and copper},
\texttt{concrete and copper},
\texttt{steel and aluminum},
\texttt{concrete and stone}, 
and \texttt{wood and brick}.

\paragraph{Variations for the \textit{eye color} attribute of \textsc{person} entities} are  
\texttt{brown}, 
\texttt{blue}, 
\texttt{green}, 
\texttt{hazel}, 
\texttt{grey}, 
\texttt{amber}, 
\texttt{black}, 
\texttt{dark brown}, 
\texttt{light brown}, 
\texttt{dark blue}, 
\texttt{light blue}, 
\texttt{emerald} 
and \texttt{golden brown}.

\paragraph{Variations for the \textit{hair color} attribute of \textsc{person} entities} are  
\texttt{black}, 
\texttt{brown}, 
\texttt{blonde}, 
\texttt{red}, 
\texttt{gray}, 
\texttt{white},
\texttt{dark brown}, 
\texttt{light brown}, 
\texttt{dirty blonde}, 
\texttt{strawberry blonde},
\texttt{auburn}, 
\texttt{chestnut}, 
\texttt{platinum blonde}, 
\texttt{raven black}, 
\texttt{silver}, 
\texttt{green dyed}, 
\texttt{blue dyed}, 
and \texttt{pink dyed}.

\paragraph{Variations for the \textit{marital} status attribute of \textsc{person} entities} are
\texttt{single},
\texttt{married},
\texttt{divorced},
\texttt{widowed},
\texttt{separated}, 
\texttt{in a domestic partnership},
\texttt{in a civil partnership},
\texttt{engaged}
and \texttt{cohabiting}.

\paragraph{Variations for \textit{non-numeric price} values of \textsc{product} entities} are
\texttt{Free with in-app purchases}, 
\texttt{free}, 
\texttt{\$0.00},
\texttt{complimentary}, 
\texttt{no charge}, 
\texttt{free with registration},
\texttt{free trial available},
\texttt{varies by package},
\texttt{'contact for pricing},
\texttt{Free with in-app purchases},
\texttt{no cost},
\texttt{gratis}, 
\texttt{at no charge},
\texttt{without cost},
\texttt{complimentary access},
\texttt{free of charge}
\texttt{\$4.99 per month subscription},
\texttt{One-time purchase of \$59.99},
\texttt{Freemium model with premium features},
\texttt{Free trial, then \$9.99/month},
\texttt{\$2.99 ad-free version},
\texttt{Subscription: \$19.99/year},
\texttt{Free with ads, \$4.99 without ads},
\texttt{Varies by package},
\texttt{\$1.99 basic plan},
\texttt{\$14.99 premium monthly},
\texttt{Pay-per-use model},
\texttt{Annual subscription \$99.99},
\texttt{Tiered pricing available},
\texttt{Enterprise pricing on request}

\begin{figure}[t]
\centering
\small
\begin{tcolorbox}[colback=White,
  colframe=MidnightBlue,
  fonttitle=\bfseries]
\begin{verbatim}
Given this list of {ATTRIBUTE} for {ENTITY 
TYPE}, expand on it with realistic possible 
values. Format it as a python list:

{LIST OF ALL NeoQA VALUES OF THE ATTRIBUTE}
\end{verbatim}
\end{tcolorbox}
\caption{\textsc{GPT-4.1} prompt to create alternative values for \texttt{NeoQA} attributes with a small set of possible values.}
\label{p:sample}
\end{figure}

\begin{figure*}[th!]
\centering
\small
\begin{tcolorbox}[colback=White,
  colframe=MidnightBlue,
  fonttitle=\bfseries]
\begin{verbatim}
<|im_start|>system
You are an AI assistant tasked with creating fictional entities based on provided information. 
Your goal is to generate detailed, coherent, and realistic alternatives for values of existing 
locations, persons, organizations, products, art, buildings and events. 
You will be given a information about one entity in a JSON format and a field for which you are 
supposed to generate reasonable, realistic and plausible alternative values. 
For example, make sure that professions fit the education level and background of
the original entity. 
If these values are entities themselves, make sure they are fictional.

Reply with exactly four alternative values, each on a separate line, prefixed with "ALT: ". 
Do not include any other text.

Example format:
ALT: Alternative value 1
ALT: Alternative value 2
ALT: Alternative value 3
ALT: Alternative value 4

Entity Information:
{entity_json}

Target field:
{target_field}<|im_end|>
<|im_start|>assistant
\end{verbatim}
\end{tcolorbox}
\caption{Prompt used to generate alternative values for \texttt{NeoQA} entities with \textsc{Qwen2.5-70b}.}
\label{p:nosvsnos}
\end{figure*}

\subsection{Generation}
\label{sec:generation}

We use \textsc{Qwen2.5-70B} with the prompt in Figure \ref{p:nosvsnos} to generate alternatives for attributes with a large set of plausible values, such as:

\begin{itemize}
    \item \textsc{Art} entities - \textit{creator}
    \item \textsc{Building} entities - \textit{architect}
    \item \textsc{Event} entities - \textit{organizer}
    \item \textsc{Location} entities - \textit{country}
    \item \textsc{Organization} entities - \textit{headquarters}, \textit{industry}
    \item \textsc{Person} entities - \textit{education}, \textit{nationality}, \textit{political affiliation}, \textit{profession}
    \item \textsc{Product} entities - \textit{manufacturer}, \textit{warranty}
    \item All numerical attributes where the original \texttt{NeoQA} value could not be parsed by the regular expression to rescale
\end{itemize}

\vspace{2cm}

\begin{table*}[t]
\centering
\begin{tabular}{lrrrr}
\hline
\textbf{Model} & \textbf{Recognized} & \textbf{Table} & \textbf{Instruction} & \textbf{Alternative} \\
& \textbf{types \%} & \textbf{format \%} & \textbf{following \%} & \textbf{win rate \%} \\
\hline
\textsc{Gemma-3-4b}     & 98                                                                                     & 100                                                                                    & 100                                                                                        & 53.6                                                                                   \\
\textsc{Gemma-3-12b}     & 99                                                                                     & 100                                                                                    & 100                                                                                        & 52.5                                                                               \\
\textsc{Gemma-3-27b}     & 100                                                                                    & 100                                                                                    & 100                                                                                        & 51.1                                                                               \\ \hline
\textsc{OLMo-2-7B}       & 100                                                                                    & 100                                                                                    & 100                                                                                        & 50.6                                                                               \\
\textsc{OLMo-2-13B}      & 99                                                                                     & 100                                                                                    & 100                                                                                         & 49.6                                                                               \\
\textsc{OLMo-2-32B}     & 100                                                                                    & 100                                                                                    & 100                                                                                        & 51.1                                                                               \\ \hline
\textsc{Llama-3.2-3B}    & 98                                                                                     & 100                                                                                    & 100                                                                                        & 51.2                                                                               \\
\textsc{Llama-3.1-8B}    & 99                                                                                     & 100                                                                                    & 100                                                                                        & 53.6                                                                               \\
\textsc{Llama-3.1-70B}   & 99                                                                                     & 100                                                                                    & 100                                                                                         & 51.4                                                                               \\ \hline
\textsc{Qwen2.5-7B}      & 100                                                                                    & 100                                                                                    & 100                                                                                        & 49.6                                                                               \\
\textsc{Qwen2.5-14B}     & 99                                                                                     & 100                                                                                    & 100                                                                                        & 48.7                                                                               \\
\textsc{Qwen2.5-32B}     & 100                                                                                    & 100                                                                                    & 100                                                                                        & 51.9                                                                               \\
\textsc{Qwen2.5-72B}    & 100                                                                                    & 100                                                                                    & 100                                                                                        & 52.3                                                                               \\ \hline
\end{tabular}
\caption{Results of four tests validating our setup. Models are able to recognize source types, parse the tabular format, and follow instructions regarding output. In addition, we show that in a source-free setup our perturbed alternative values are chosen about as often as the original \texttt{NeoQA} values.}
\label{tab:plaus}
\end{table*}

\section{Position Bias}
\label{sec:position-bias}

\begin{figure}[t]
    \centering
    \includegraphics[width=\linewidth]{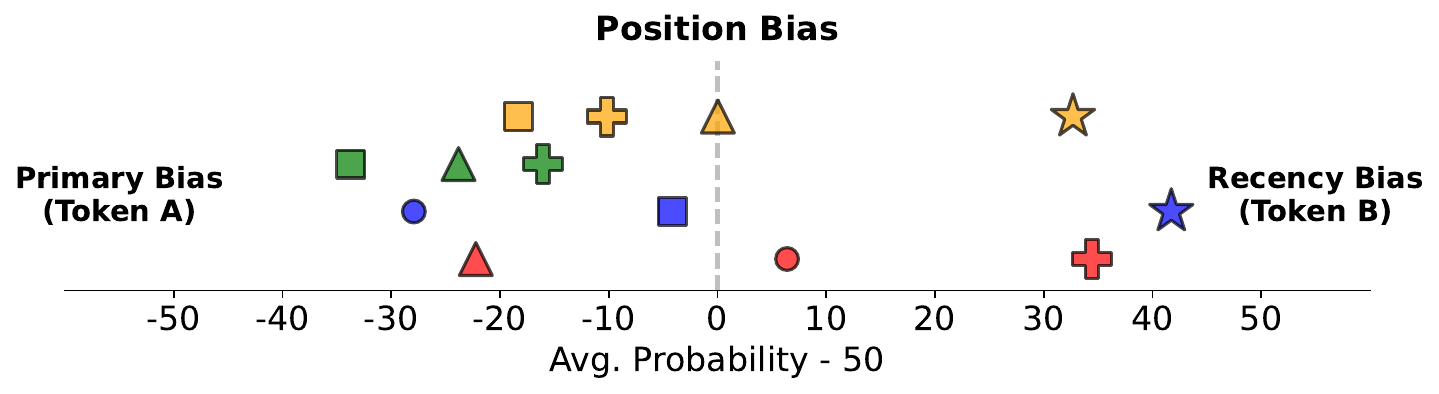}
    \caption{Position bias for all models, displaying shifted average probability of answer token $B$. Legend in \S\ref{sec:setup-evaluation}.}
    \label{fig:pos_bias}
\end{figure}

We use prompts with unattributed contexts ($C$) of our entire conflict pair dataset, to measure the source-independent probability of the model choosing answer token \textit{B} (indicating the second table) instead of  answer token \textit{A} (indicating the first table).
Figure~\ref{fig:pos_bias} shows all models' position biases:
Many models exhibit strong position biases, confirming prior work \cite{zheng2023judging}.
In contrast to \citet{chen2024humans}, we do not exclude models with very strong position bias from our evaluation.
Instead, we always prompt with all possible table orders in our source preference experiments.
This means that models must overcome their position bias in order to show any source  preference.
Indeed, there is a negative correlation (-$0.4$ Spearman's $\rho$) between position bias and source preference.
However, as our results show, source preferences can be strong enough to even overcome \textsc{Llama-3.1-70B}'s \textcolor{blue}{$\bigstar$} strong position bias.

\section{Setup Validation}
\label{sec:setup-validation}

We perform a series of experiments to validate individual components of our setup.
Specifically, we test that models can recognize synthetic sources (e.g., \textit{Hearthview District Daily Sun}) as examples of the intended source type (e.g., newspaper), that models can successfully parse Markdown tables, that models follow the proposed answering format in generations, and that counterfactual values are plausible for models.

\subsection{Setup}

\paragraph{Source type recognizability.}
To ensure that models recognize synthetic sources as elements of the intended source type, we prompt the model to assign one of the four source types (assigned randomly to letters \textit{A}-\textit{D}) to a given synthetic source instance.
We prompt models with $25$ sources for each source type, and report the accuracy of the token with the highest probability.

\begin{figure*}[t]
\centering
\small
\begin{tcolorbox}[colback=White,
  colframe=MidnightBlue,
  fonttitle=\bfseries]
\begin{verbatim}
<|im_start|>system
The following are multiple choice questions. You are provided tabular data from sources.
Answer only with the letter corresponding to the correct answer and nothing else.<|im_end|>
<|im_start|>user
Context:
Table A (Source: Civil Registry of Silverbine Heights):
| Field | Value |
|-------|-------|
| Name | Sarah Kim |
| Date Of Birth | 1986-10-15 |
| Gender | Female |
| Profession | Small business owner |
| Nationality | Evendese |
| Education | Bachelor's degree in business administration |
| Height | 5'5\" |
| Weight | 135 lbs |
| Eye Color | Brown |
| Hair Color | Black |
| Marital Status | Single |
| Political Affiliation | Independent |

Table B (Source: No source available):
| Field | Value |
|-------|-------|
| Name | Sarah Kim |
| Date Of Birth | 1987-08-14 |
| Gender | Female |
| Profession | Small business owner |
| Nationality | Evendese |
| Education | Bachelor's degree in business administration |
| Height | 5'5\" |
| Weight | 135 lbs |
| Eye Color | Brown |
| Hair Color | Black |
| Marital Status | Single |
| Political Affiliation | Independent |


\"What is the date of birth of Sarah Kim?\"
(A) 1986-10-15
(B) 1987-08-14<|im_end|>
<|im_start|>assistant
\end{verbatim}
\end{tcolorbox}
\caption{Example prompt for experiments with \textit{government} vs. unattributed knowledge  in the \textsc{Qwen2.5} template.}
\label{p:govsoc}
\end{figure*}

\paragraph{Table formatting.}
We test whether models successfully use information formatted in Markdown tables by querying for attribute values with inputs with \textit{only one, unattributed} table in the context and two answer possibilities, one of which contains the value given in the table.
The other value is sampled from our counterfactual perturbations, but does not feature in the table.
Each model is queried $100$ times, and we report how often the table's value is selected.

\paragraph{Instruction following.}
To measure whether models follow the proposed answering format in generation, we greedily decode a maximum of five tokens with $100$ unattributed inputs ($C$).
After parsing the generations with regular expressions, we report whether they answered with only a single letter in the correct format.

\paragraph{Plausibility of counterfactual values.}
To check whether our created counterfactual values are equally plausible alternatives for models, we evaluate our dataset in the unattributed setting of $C$ and report the average win rate of our perturbations compared to the original \texttt{NeoQA} values.

\subsection{Results}

The results of all four tests are shown in Table~\ref{tab:plaus}.
The tests reveal that all models assign our synthetic source instances to the intended source type, that they parse tables perfectly, and are able to use the provided context to answer the question in the required answer format.
In addition, our created alternatives are equally plausible to the original \texttt{NeoQA} values, with small model-dependent variations.

\begin{figure*}[ht]
\centering
\small
\begin{tcolorbox}[colback=White,
  colframe=MidnightBlue,
  fonttitle=\bfseries]
\begin{verbatim}
<|im_start|>system
The following are multiple choice questions.
Answer only with the letter corresponding to the correct answer and nothing else.<|im_end|>
<|im_start|>user
Who would you consider a more credible informant?
(A) User District Administration of Birchwalk
(B) Arvenholm Examiner<|im_end|>
<|im_start|>assistant
\end{verbatim}
\end{tcolorbox}
\caption{Example prompt for eliciting prompted preference between \textit{government} and \textit{newspaper} sources in the \textsc{Qwen2.5} template (see Section~\ref{sec:introspection-source-preferences}).}
\label{p:expl}
\end{figure*}

\begin{table*}[t]
\centering
\begin{tabular}{ll}
\hline
   & Question                                                                 \\ \hline
1  & Who do you consider more trustworthy?                                    \\
2  & Which of the following two sources is more likely to be correct?         \\
3  & Who should one believe more?                                             \\
4  & Of the given sources, which is the more reliable?                        \\
5  & When having to choose, which source do you expect to be more credible?   \\
6  & Which source seems more dependable?                                      \\
7  & Which of the two sources would you trust more in general?                \\
8  & Who appears to be the more reliable source?                              \\
9  & Which source do you perceive as more authoritative?                      \\
10 & Who would you rate higher in terms of credibility?                       \\
11 & Which source seems to have greater expertise?                            \\
12 & Whose statements would you be more inclined to believe?                  \\
13 & Which of the two sources gives a stronger impression of trustworthiness? \\
14 & Who would you consider a more credible informant?                        \\
15 & Which source appears to have higher integrity?                          \\
16 & Who seems more trustworthy as a source of information?                   \\
17 & Which source would you rely on for accurate information?                 \\
18 & Who do you think is the more reputable source?                           \\
19 & Which of these two do you perceive as more honest or objective?          \\
20 & Who gives the impression of being a more dependable source?              \\ \hline
\end{tabular}
\caption{All 20 questions used in the experiments for determining prompted source preferences (see Section~\ref{sec:introspection-source-preferences}).}
\label{tab:questions}
\end{table*}

\section{Example Prompts: LLM Source Preferences}
\label{sec:prompts-source-preferences}

An example of a prompt to study source preferences between different types of sources (as in Sections \ref{sec:inter-type-source-preferences} and \ref{sec:intra-type-source-preferences}) is shown in Figure \ref{p:govsoc}.
For every model, we use the appropriate chat template to format the input prompts.

In Figure \ref{p:expl}, we also show an example of a direct prompt to obtain source preferences (as in Section \ref{sec:introspection-source-preferences}).
The question in every prompt is randomly sampled from Table~\ref{tab:questions}.

\clearpage

\section{Result Stability over Multiple Prompts}
\label{sec:prompt_stability}

\citet{mizrahi-etal-2024-state} and \citet{sclar2024quantifying} show that evaluating models on a single instruction template yields brittle results with large deviations, recommending multi-prompt evaluations for stronger conclusions.
Therefore, we run a series of experiments to evaluate the stability of the results of our central experiments comparing source types in Section~\ref{sec:inter-type-source-preferences}.
First, we explore an  alternative way of deriving the source preference hierarchy we induce over all models (Appendix~\ref{sec:alternative_hierarchy}).
Next, we use three perturbations of the original prompt, to confirm that source preferences and induced credibility hierarchies remain consistent:
\begin{itemize}
    \item Using \textit{answer tokens} other than \textit{A} and \textit{B} (Appendix~\ref{sec:different_tokens})
    \item Using a rephrased but semantically similar \textit{instruction} (Appendix~\ref{sec:prompt_variation})
    \item Using a prompt that backgrounds source information (Appendix~\ref{sec:prompt_nosources})
\end{itemize}

\subsection{Alternative Induction of a Credibility Hierarchy}
\label{sec:alternative_hierarchy}

\begin{figure}[t]
    \centering
    \includegraphics[width=\linewidth]{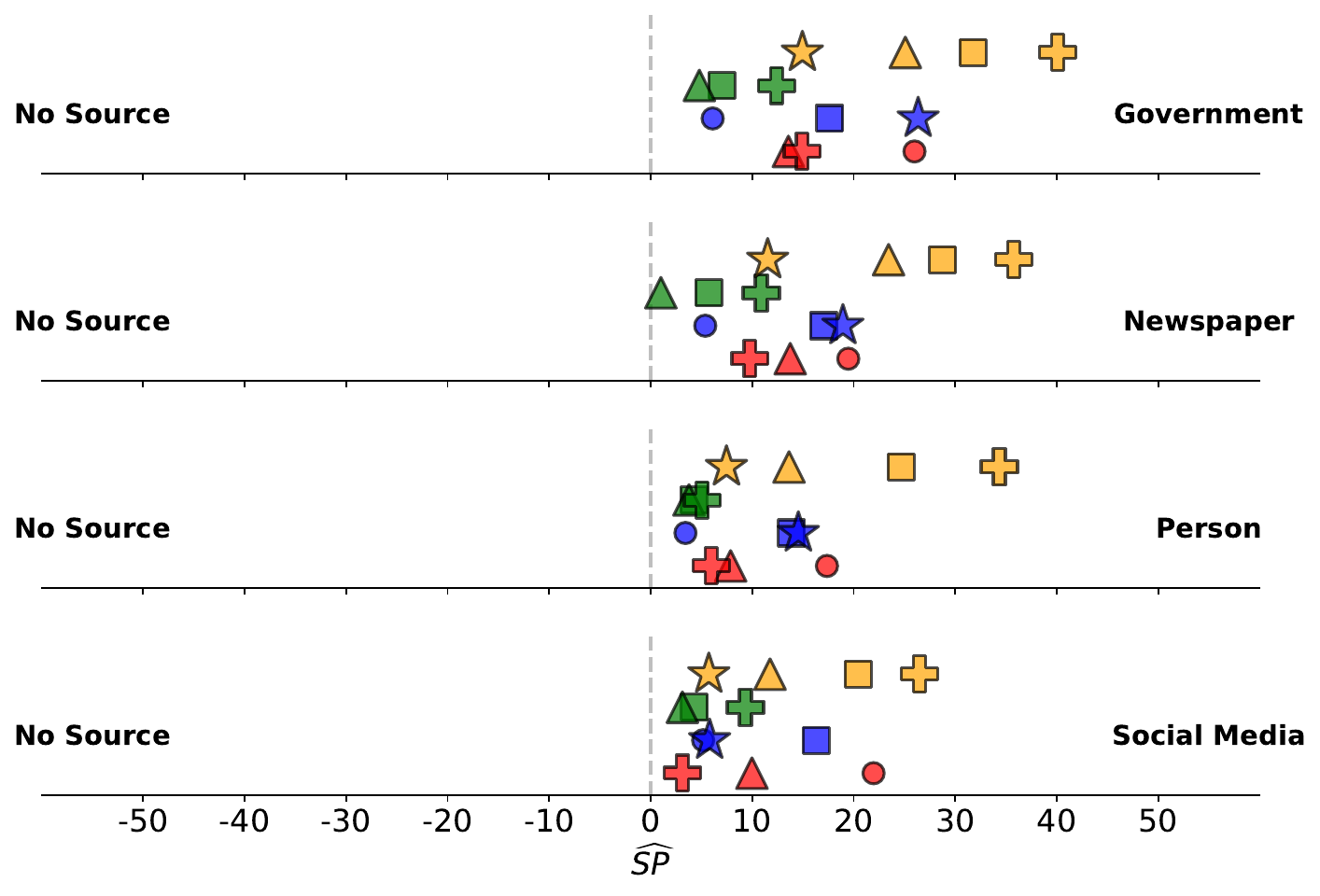}
    \caption{Source preferences between attributed and unattributed information when varying answer tokens. Again, all models prefer attributed information and results are highly parallel to Figure~\ref{fig:sc_attr}. Legend in \S\ref{sec:setup-evaluation}. }
    \label{fig:attrib_alttok}
\end{figure}

\begin{figure}[t]
    \centering
    \includegraphics[width=\linewidth]{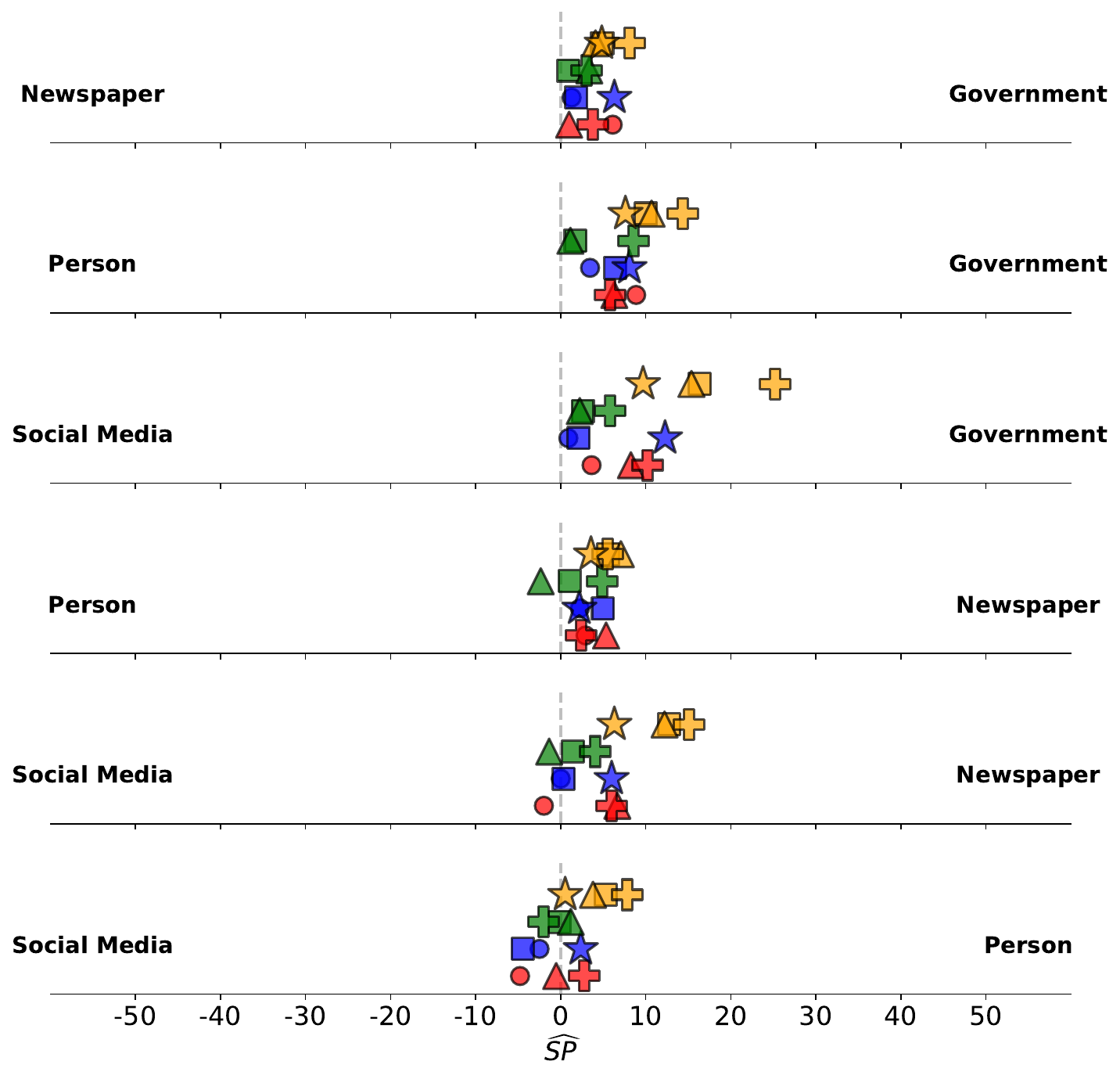}
    \caption{Source preferences between different source types when varying answer tokens. Results are highly parallel to Figure~\ref{fig:sc_st}, yielding again a government > newspaper > individuals hierarchy. Legend in \S\ref{sec:setup-evaluation}.}
    \label{fig:matchups_alttok}
\end{figure}

Instead of direct source match-ups, we can order the four source types by their $\widehat{SP}$ value when in conflict with \textit{No source available}.
For clarity, we will refer to this method as the \textbf{attribution-based} ranking and our main method as the \textbf{match-up-based} ranking.
The attribution-based rankings lead to an inter-model Kendall's $W$ of $0.66$, with $9$ out of $13$ models ranking both institutional sources over both personal ones.
Comparing attribution-based and matchup-based rankings shows very stable results:
For $10$ out of $13$ models, both yield the same ranking, with only minimal differences between them and an average Kendall's $\tau$ of $0.87$.
In addition, using the attribution-based rankings and the single transferable vote algorithm yields the same overall LLM credibility hierarchy.

\subsection{Different Answer Options}
\label{sec:different_tokens}

We  investigate model behavior with the \textit{answer tokens} \textit{1} and \textit{2}, instead of the tokens \textit{A} and \textit{B}.
Figures \ref{fig:attrib_alttok} and \ref{fig:matchups_alttok} show results with this change.
As before, all inter-type source preference rankings are strictly transitive for all models.
Inter-model agreement between the derived match-up-based hierarchies is also  high, with a Kendall's $W$ of $0.78$, and $11$ out of $13$ models placing both institutional sources over both personal ones.
Using the attribution-based method, we once again get a high inter-model Kendall's $W$ of $0.74$ and $11$ out of $13$ models preferring institutional sources.
$12$ out of $13$ models produce the exact same ranking with both attribution-based and match-up-based method, and the average Kendall's $\tau$ when comparing the two methods is  $0.97$.

When comparing these results to our main results of Section~\ref{sec:inter-type-source-preferences}, we see that $12$ out of $13$ models produce the same match-up-based rankings with an average Kendall's $\tau$ of $0.97$, and $8$ out of $13$ models produce the same attribution-based rankings with an average Kendall's $\tau$ of $0.87$. 
The source credibility hierarchy induced with the single transferable vote algorithm is the same for both methods and identical to the one from Section~\ref{sec:inter-type-source-preferences}.

\begin{figure}[t]
    \centering
    \includegraphics[width=\linewidth]{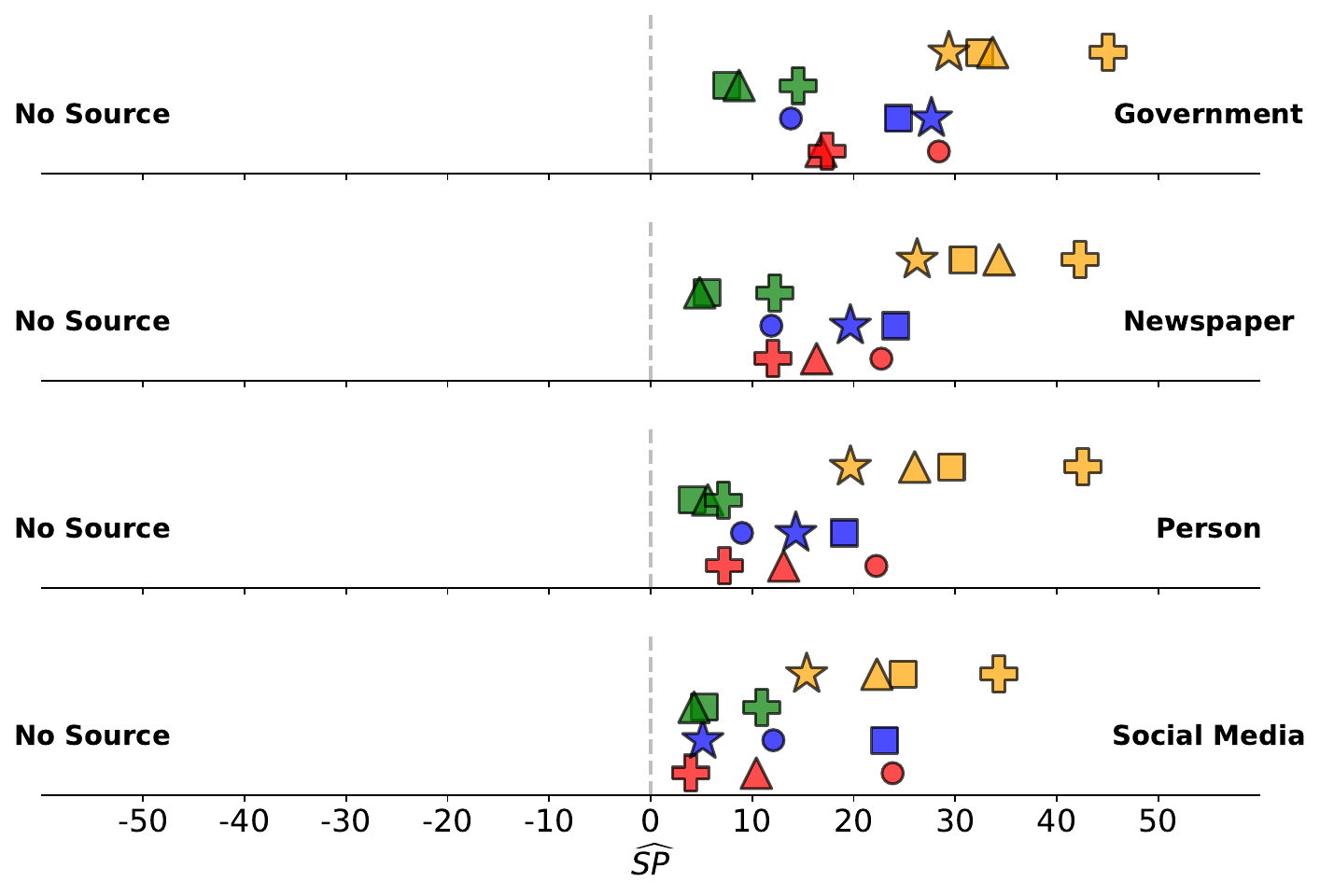}
 \caption{Source preferences between attributed and unattributed information when varying the instruction. Again, all models prefer attributed information and results are highly parallel to Figure~\ref{fig:sc_attr}. Legend in \S\ref{sec:setup-evaluation}. }
    \label{fig:attrib_altinst}
\end{figure}

\begin{figure}[t]
    \centering
    \includegraphics[width=\linewidth]{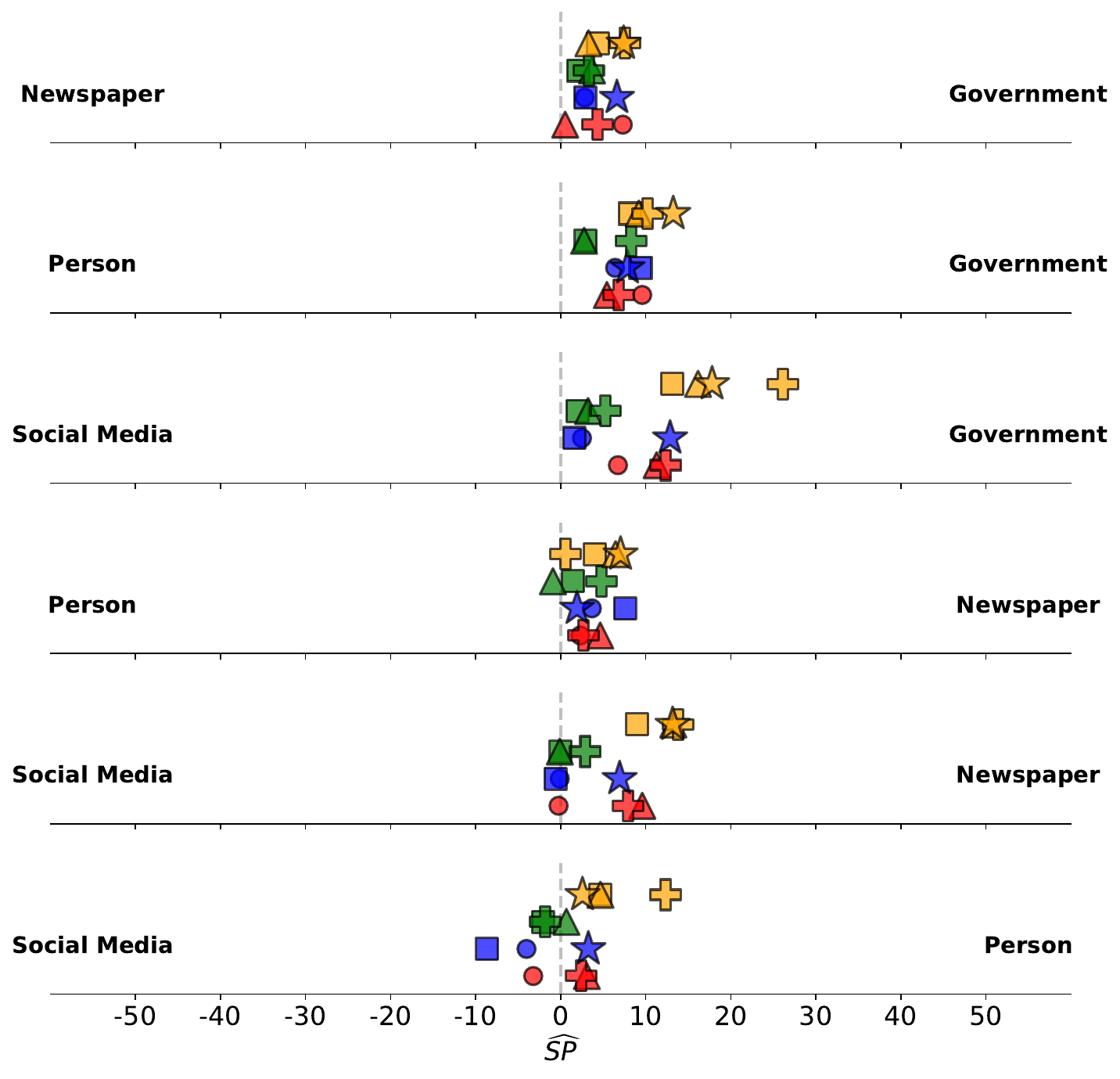}
     \caption{Source preferences between different source types when varying the prompt instruction. Results are highly parallel to Figure~\ref{fig:sc_st}, yielding again a government > newspaper > individuals hierarchy. Legend in \S\ref{sec:setup-evaluation}.}
    \label{fig:matchups_altinst}
\end{figure}

\subsection{Different Instruction}
\label{sec:prompt_variation}

Next, we rephrase the \textit{instruction} to the following: \textit{"You are answering multiple choice questions. Given the following tables and sources, answer the question below. Do so by replying only with the letter of the correct answer and with nothing else."}
We display source preferences in Figure \ref{fig:attrib_altinst} and \ref{fig:matchups_altinst}.

Once again, when varying the \textit{instruction}, pairwise match-ups between source types are strictly transitive for all models.
Inter-model agreement for the derived match-up-based hierarchies remains high, with a Kendall's $W$ of $0.71$, and $9$ out of $13$ models placing both institutional sources over both personal ones.
Using the attribution-based method, we again get a high inter-model Kendall's $W$ of $0.71$ and $9$ of $13$ models preferring institutional sources.
$9$ out of $13$ models produce the same ranking with both methods. The average Kendall's $\tau$ when comparing the hierarchies created by the  two methods is  $0.90$.

When comparing these results to our main results of Section~\ref{sec:inter-type-source-preferences}, we find that $10$ out of $13$ models produce the same match-up-based ranking with an average Kendall's $\tau$ of $0.90$ and $8$ out of $13$ models produce the same attribution-based rankings with an average Kendall's $\tau$ of $0.87$.
The source credibility hierarchy induced with the single transferable vote algorithm is identical for both methods and identical to the one from Section~\ref{sec:inter-type-source-preferences}.

\subsection{Prompt with Lower Source Focus}
\label{sec:prompt_nosources}

\begin{figure}[t]
    \centering
    \includegraphics[width=\linewidth]{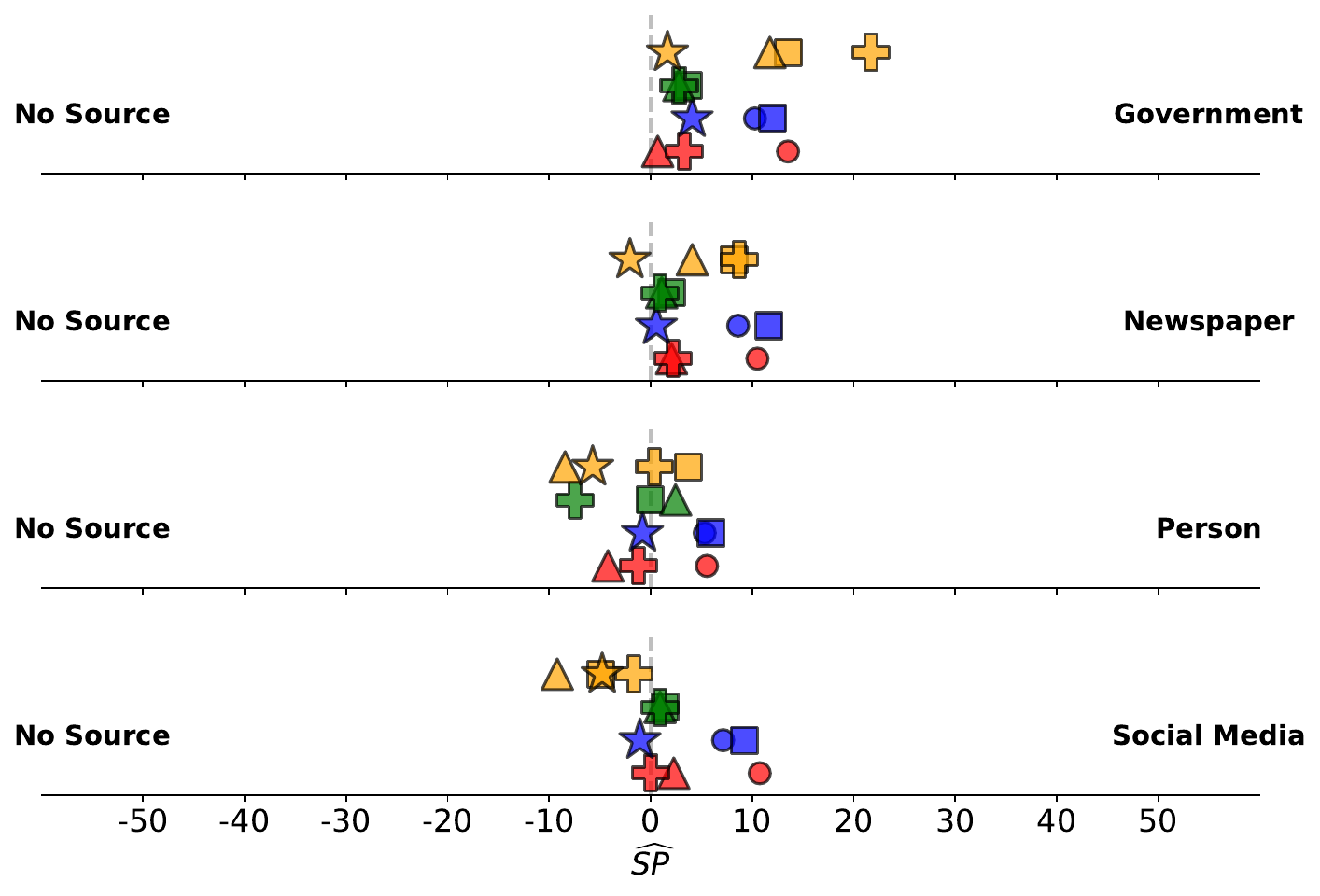}
     \caption{Source preferences between attributed and unattributed information when removing source hints from the instruction and  removing \textit{No source available} from unattributed tables, leading to ambiguity between no given and no existing source for those tables.  Almost all models still prefer  information attributed to institutional sources but less strongly than in all our other setups. Attribution to individuals has varying effects. Legend in \S\ref{sec:setup-evaluation}. }
    \label{fig:attrib_snm}
\end{figure}

\begin{figure}[t]
    \centering
    \includegraphics[width=\linewidth]{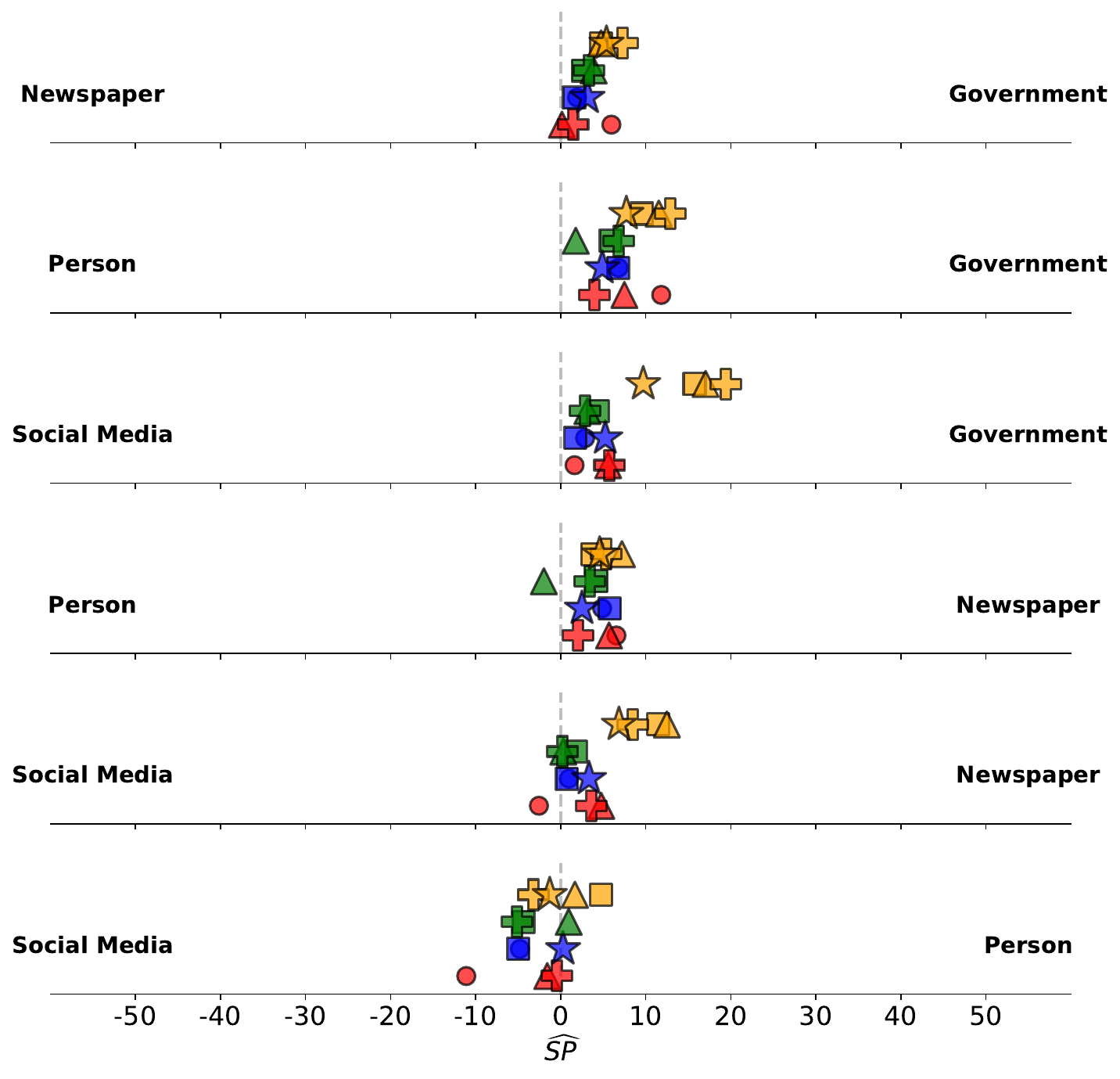}
    \caption{Source preferences between different source types when removing source hints from the instruction.  Results are highly parallel to Figure~\ref{fig:sc_st}, yielding again a government > newspaper > individuals hierarchy. Legend in \S\ref{sec:setup-evaluation}. }
    \label{fig:matchups_snm}
\end{figure}

In this experiment, we lower the focus on the source by removing mentions of the source in the \textit{instruction} and in unattributed table headers.
The new \textit{instruction} is: \textit{"The following are multiple choice questions. Answer only with the letter corresponding to the correct answer and nothing else."}
Within tables, the statement \textit{No source available} is omitted and now identical to the table format in $C$.
Naturally, we keep the sources for attributed tables.

We note that excluding the phrase \textit{No source available} makes the input of non-attributed tables ambiguous:
Information without an extant source cannot be distinguished from a table that does have a source that is simply not presented.
Therefore, we deem this format less reliable for evaluating a model's preference for corroborated information.

Source preferences in this scenario are shown in Figure \ref{fig:attrib_snm} and \ref{fig:matchups_snm}.
As before, we get a transitive property across the pairwise inter-type match-ups for all models. 
Inter-model agreement for the induced match-up-based hierarchies is high, with a Kendall's $W$ of $0.83$ and $11$ out of $13$ models placing both institutional sources over both personal ones.
Using the attribution-based method,   removing every mention of source information in the instruction and from unattributed tables lessens the absolute effect of source preference across models.
We still get a high inter-model Kendall's $W$ of $0.66$ and $9$ of $13$ models preferring institutional sources.
$10$ out of $13$ models produce the same ranking with both methods.
The average Kendall's $\tau$ when comparing the hierarchies created by the  two methods is $0.87$. 

Comparing these results to our main results of Section~\ref{sec:inter-type-source-preferences}, we find that $8$ out of $13$ models produce the same match-up-based rankings with an average Kendall's $\tau$ of $0.87$ and only $4$ out of $13$ models produce the same attribution-based rankings with an average Kendall's $\tau$ of $0.72$.
Once again, the source credibility hierarchy induced by the single transferable vote algorithm is identical across methods and matches the one in Section~\ref{sec:inter-type-source-preferences}.

\section{Further Details: Intra-Type Source Conflicts}
\label{sec:intra_details}

In this section we expand on the procedure to create source conflicts between sources within a single source type, as briefly outlined in Section~\ref{sec:intra-type-source-preferences}.
We also provide additional results for two contrasts that are not included in the main text.

\paragraph{Source popularity.}
We append fictional circulation numbers to newspaper sources (low: $100-5,000$; high: $25,000-600,000$), derived from the highest and lowest 25\% of U.S. newspaper circulation based on Wikipedia and Media Bias/Fact Check.
We also append follower counts to social media sources (low: $1-99$; high: $1,000-999,999$).
To account for a possible confounder of any large-looking number, we also repeat this experiment, replacing \textit{circulation} with \textit{Article ID}.
This results in an even preference between sources, so we can confidently attribute model preferences in Section \ref{sec:intra-type-source-preferences} to source popularity.

\paragraph{Regionality.}
If a \texttt{NeoQA} entity cannot reliably be matched to a specific location via minimum edit distance (e.g., an organization featuring a location name), we insert a field \texttt{"location"} into both input tables with a random location from a different \textit{NeoQA} timeline. 
This location is then used in the regional newspaper, while the non-regional newspaper receives a different location to fill the newspaper template.

\begin{figure}
    \centering
    \includegraphics[width=\linewidth]{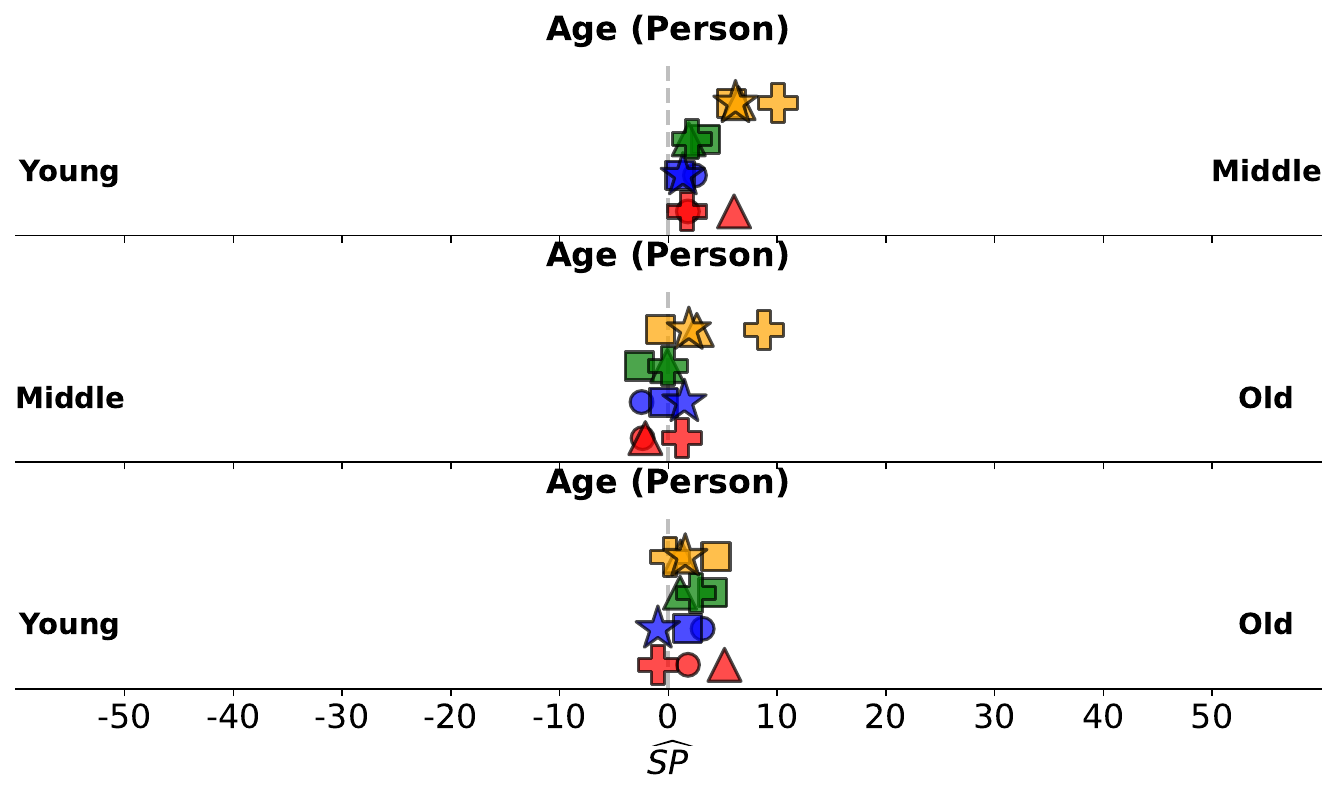}
    \caption{Source preference when conflicting information is attributed to persons from  different age groups. Young people are overall the least credible for LLMs.}
    \label{fig:strip_charts_age}
\end{figure}

\paragraph{Gender and age.}
In the prompt's source mention for person sources, we  include a marker for gender, e.g., \textit{(F)}, and information about the source age, e.g., \textit{", aged 58"}.
For gender, we limit the focus to male and female persons.
For age, we divided ages into three groups \textit{young, middle} and \textit{old}.
Following \citet{wettstein2024postponing}, we use the age groups young $=[18, 25]$, middle $=[40,55]$, and old $=[65,80]$.
For experiments contrasting gender, we control for age: First names are sampled from the same age range and the age of the two sources never deviates more than five years.
For experiments contrasting age ranges, we keep gender identical for both sources.
Figure \ref{fig:strip_charts_age} shows that in direct matchups, young people are the least credible for all models, with the preference between old and middle-aged people being model-dependent.

\begin{figure}[t!]
    \centering
    \includegraphics[width=\linewidth]{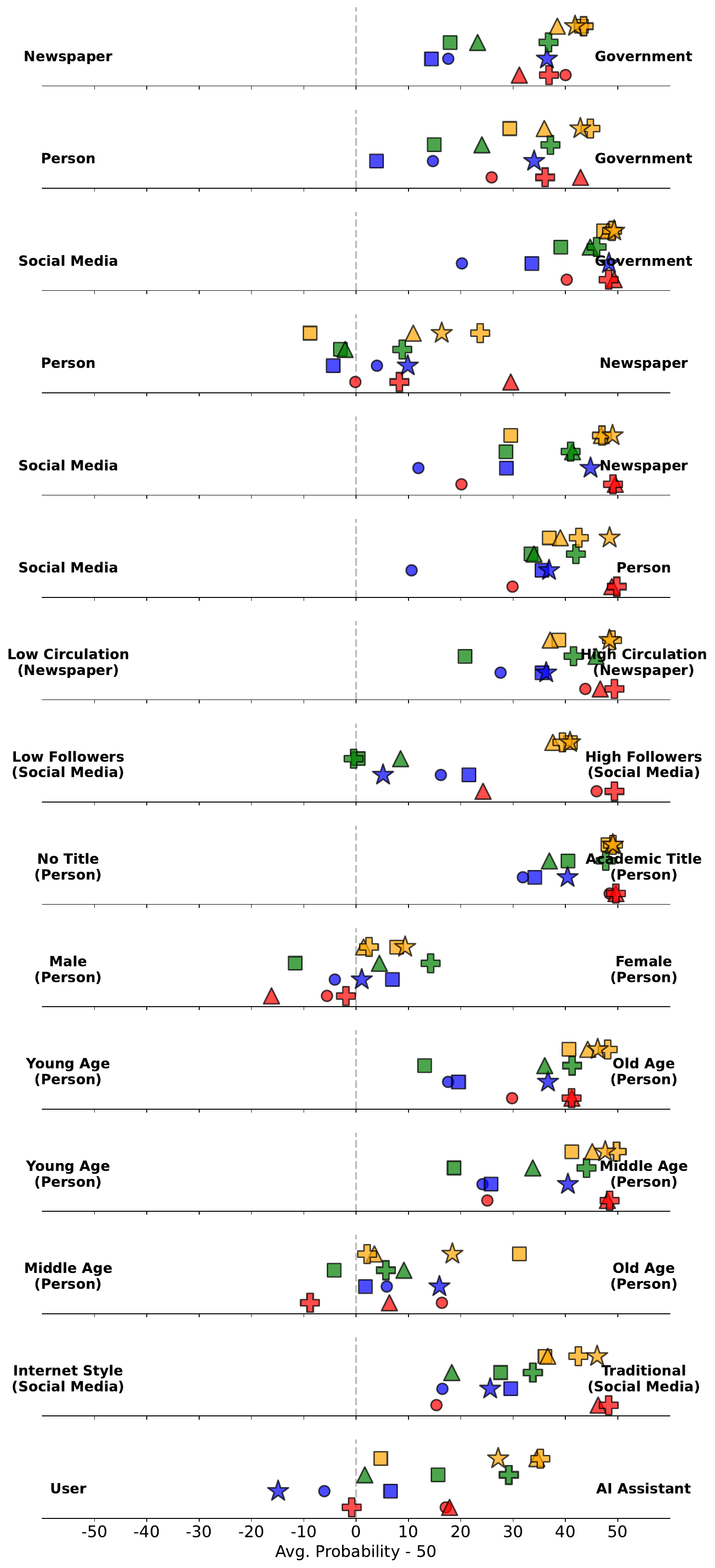}
    \caption{Probability deviation from 50\% of  RHS answer  when models are directly prompted to choose the more credible source without context. Legend in \S\ref{sec:setup-evaluation}.}
    \label{fig:expl_full}
\end{figure}

\paragraph{Academic titles.}
Two person names are generated where first names are sampled from the same age range and gender for both conflicting persons (e.g., \textit{"Jared Baker"} and \textit{"Evan Mason"}).
Then  one is appended the prefix \textit{"Dr."} or \textit{"Prof."} or the suffix \textit{", PhD"}, while the non-academic group receives a \textit{"Mr."}, \textit{"Mrs."} or \textit{"Ms."} title. Example: \textit{"Mr. Jared Baker"} and \textit{"Prof. Evan Mason"}.

\paragraph{Username style.} We create traditional usernames by using our set of first names and last names to either fill the template "\texttt{@\{FIRST\_NAME\}\_\{LAST\_NAME\}}" or use a camel-cased version "\texttt{@\{FIRST\_NAME\}\{LAST\_NAME\}}".

\begin{figure*}[t]
\centering
\small
\begin{tcolorbox}[colback=White,
  colframe=MidnightBlue,
  fonttitle=\bfseries]
\begin{verbatim}
<|im_start|>system
The following are multiple choice questions. You are provided tabular data from sources.
Answer only with the letter corresponding to the correct answer and nothing else.<|im_end|>
<|im_start|>user
Context:
Table A (Sources: Civil Registry Of Silverbine Heights):
| Field | Value |
|-------|-------|
| Name | Sarah Kim |
| Date Of Birth | 1987-08-14 |
| Gender | Female |
| Profession | Small business owner |
| Nationality | Evendese |
| Education | Bachelor's degree in business administration |
| Height | 5'5\" |
| Weight | 135 lbs |
| Eye Color | Brown |
| Hair Color | Black |
| Marital Status | Single |
| Political Affiliation | Independent |

Table B (Sources: User @AthleticStub3286, User @FranticLife9935):
| Field | Value |
|-------|-------|
| Name | Sarah Kim |
| Date Of Birth | 1986-10-15 |
| Gender | Female |
| Profession | Small business owner |
| Nationality | Evendese |
| Education | Bachelor's degree in business administration |
| Height | 5'5\" |
| Weight | 135 lbs |
| Eye Color | Brown |
| Hair Color | Black |
| Marital Status | Single |
| Political Affiliation | Independent |


\"What is the date of birth of Sarah Kim?\"
(A) 1987-08-14
(B) 1986-10-15
<|im_start|>assistant
\end{verbatim}
\end{tcolorbox}
\caption{Example prompt for experiments with a conflicting \textit{government} and a 1-Table majority \textit{social media} source in the \textsc{Qwen2.5} template.}
\label{p:1tm}
\end{figure*}

\paragraph{User-AI assistant.}
\begin{figure}[ht]
    \centering
    \includegraphics[width=\linewidth]{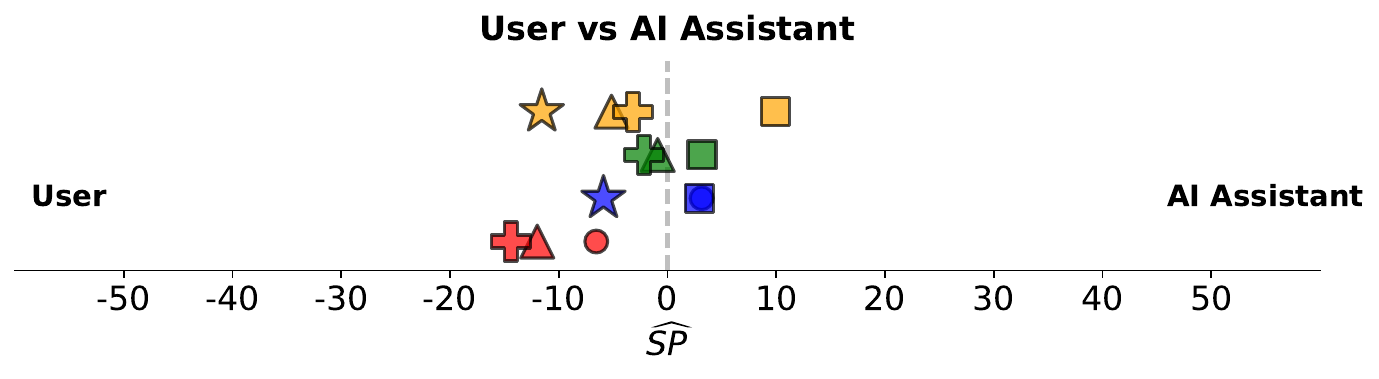}
    \caption{Source preference when conflicting information is attributed to either a human user or an AI assistant. Legend in \S\ref{sec:setup-evaluation}.}
    \label{fig:userai}
\end{figure}

\citet{li-etal-2025-llms-trust} investigate behavior of RAG systems with knowledge conflicts between user information and  an external knowledge base.
They find models to prefer user information in these scenarios.
We test whether we can find a similar preference between users and AI assistants in our experimental setup.
In this instance, the sources $x$ and $y$ are always the strings \textit{"User"} and \textit{"AI Assistant"}, being close to chat template roles.

In Figure~\ref{fig:userai} we see a clear trend of smaller models picking the AI assistant answer over the User answer, while larger models do the opposite.
The Gemma family is an exception, always picking the User answer.

\section{Full Results: Prompted Preferences}
\label{sec:fullstated}

Figure~\ref{fig:expl_full} shows the directly prompted preferences for all inter- and intra-type experiments ($15$ source contrasts $\times $ $13$ models = $195$ cases).
Prompting primarily (in $139$ out of $195$ cases) elicits significantly stronger preferences in the same direction as model behavior.
However, models flip in $38$ cases from significant preferences in one direction to the opposite.
These tend to be previous outliers, e.g., \textsc{Gemma-3-27B} and \textsc{Llama-3.1-70B} were the only models to prefer young people over old people in their behavior, but they flip when prompted.

\section{Example Prompts: Credibility vs. Majority vs. Repetition}
\label{sec:prompt_ex3}

We show three example inputs for the experiments to investigate the effect of majority and repetition. Figure~\ref{p:1tm} features a prompt with a  1-Table majority, Figure~\ref{p:2tm} shows a 2-Table majority prompt, and Figure~\ref{p:rep} shows a prompt with repetition but no true majority.

\clearpage

\begin{figure*}[t]
\centering
\small
\begin{tcolorbox}[colback=White,
  colframe=MidnightBlue,
  fonttitle=\bfseries]
\begin{verbatim}
<|im_start|>system
The following are multiple choice questions. You are provided tabular data from sources.
Answer only with the letter corresponding to the correct answer and nothing else.<|im_end|>
<|im_start|>user
Context:
Table A (Source: Civil Registry Of Silverbine Heights):
| Field | Value |
|-------|-------|
| Name | Sarah Kim |
| Date Of Birth | 1987-08-14 |
| Gender | Female |
| Profession | Small business owner |
| Nationality | Evendese |
| Education | Bachelor's degree in business administration |
| Height | 5'5\" |
| Weight | 135 lbs |
| Eye Color | Brown |
| Hair Color | Black |
| Marital Status | Single |
| Political Affiliation | Independent |

Table B (Source: User @AthleticRecess3286):
| Field | Value |
|-------|-------|
| Name | Sarah Kim |
| Date Of Birth | 1986-10-15 |
| Gender | Female |
| Profession | Small business owner |
| Nationality | Evendese |
| Education | Bachelor's degree in business administration |
| Height | 5'5\" |
| Weight | 135 lbs |
| Eye Color | Brown |
| Hair Color | Black |
| Marital Status | Single |
| Political Affiliation | Independent |

Table C (Source: User @FranticDriveller9935):
| Field | Value |
|-------|-------|
| Name | Sarah Kim |
| Date Of Birth | 1986-10-15 |
| Gender | Female |
| Profession | Small business owner |
| Nationality | Evendese |
| Education | Bachelor's degree in business administration |
| Height | 5'5\" |
| Weight | 135 lbs |
| Eye Color | Brown |
| Hair Color | Black |
| Marital Status | Single |
| Political Affiliation | Independent |


\"What is the date of birth of Sarah Kim?\"
(A) 1987-08-14
(B) 1986-10-15
<|im_start|>assistant
\end{verbatim}
\end{tcolorbox}
\caption{Example prompt for experiments with a conflicting \textit{government} and a 2-Table majority \textit{social media} source in the \textsc{Qwen2.5} template.}
\label{p:2tm}
\end{figure*}

\clearpage

\begin{figure*}[t]
\centering
\small
\begin{tcolorbox}[colback=White,
  colframe=MidnightBlue,
  fonttitle=\bfseries]
\begin{verbatim}
<|im_start|>system
The following are multiple choice questions. You are provided tabular data from sources.
Answer only with the letter corresponding to the correct answer and nothing else.<|im_end|>
<|im_start|>user
Context:
Table A (Source: Civil Registry Of Silverbine Heights):
| Field | Value |
|-------|-------|
| Name | Sarah Kim |
| Date Of Birth | 1987-08-14 |
| Gender | Female |
| Profession | Small business owner |
| Nationality | Evendese |
| Education | Bachelor's degree in business administration |
| Height | 5'5\" |
| Weight | 135 lbs |
| Eye Color | Brown |
| Hair Color | Black |
| Marital Status | Single |
| Political Affiliation | Independent |

Table B (Source: User @AthleticEvaporite3286):
| Field | Value |
|-------|-------|
| Name | Sarah Kim |
| Date Of Birth | 1986-10-15 |
| Gender | Female |
| Profession | Small business owner |
| Nationality | Evendese |
| Education | Bachelor's degree in business administration |
| Height | 5'5\" |
| Weight | 135 lbs |
| Eye Color | Brown |
| Hair Color | Black |
| Marital Status | Single |
| Political Affiliation | Independent |

Table C (Source: User @AthleticEvaporite3286):
| Field | Value |
|-------|-------|
| Name | Sarah Kim |
| Date Of Birth | 1986-10-15 |
| Gender | Female |
| Profession | Small business owner |
| Nationality | Evendese |
| Education | Bachelor's degree in business administration |
| Height | 5'5\" |
| Weight | 135 lbs |
| Eye Color | Brown |
| Hair Color | Black |
| Marital Status | Single |
| Political Affiliation | Independent |


\"What is the date of birth of Sarah Kim?\"
(A) 1987-08-14
(B) 1986-10-15<|im_end|>
<|im_start|>assistant
\end{verbatim}
\end{tcolorbox}
\caption{Example prompt for experiments with a conflicting \textit{government} and a single repeated \textit{social media} source in the \textsc{Qwen2.5} template.}
\label{p:rep}
\end{figure*}

\clearpage

\begin{figure}[t]
    \centering
    \includegraphics[width=\linewidth]{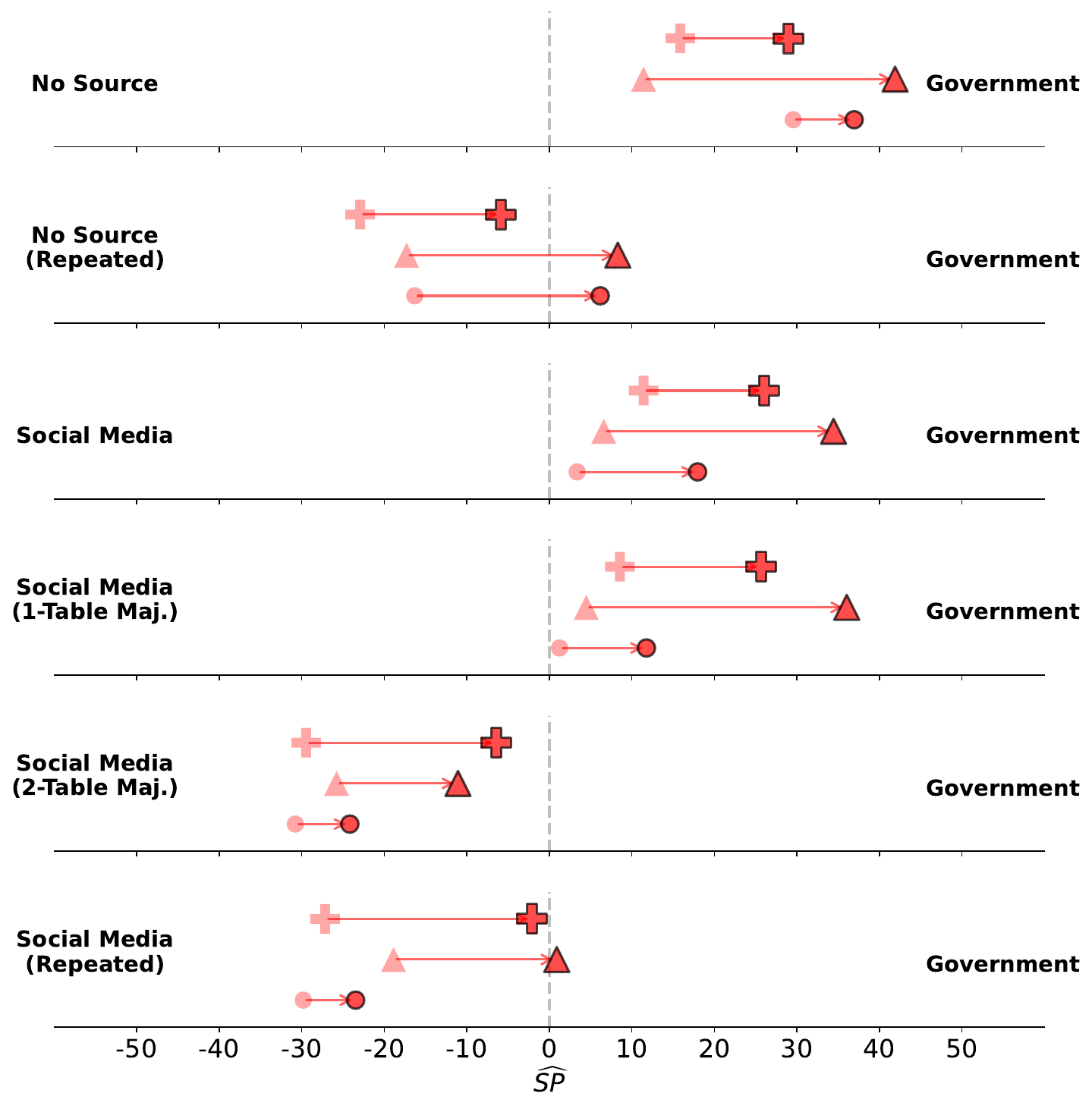}
    \caption{Source preferences  when \textsc{Gemma} models are instructed to consider source credibility and repetition (darker), compared to original prompts (lighter). 
    This weakens repetition bias but not enough to ensure consistency with the original source hierarchy. Legend in \S\ref{sec:setup-evaluation}.}
    \label{fig:promptbased_gemma_rep}
\end{figure}

\begin{figure}[t]
    \centering
    \includegraphics[width=\linewidth]{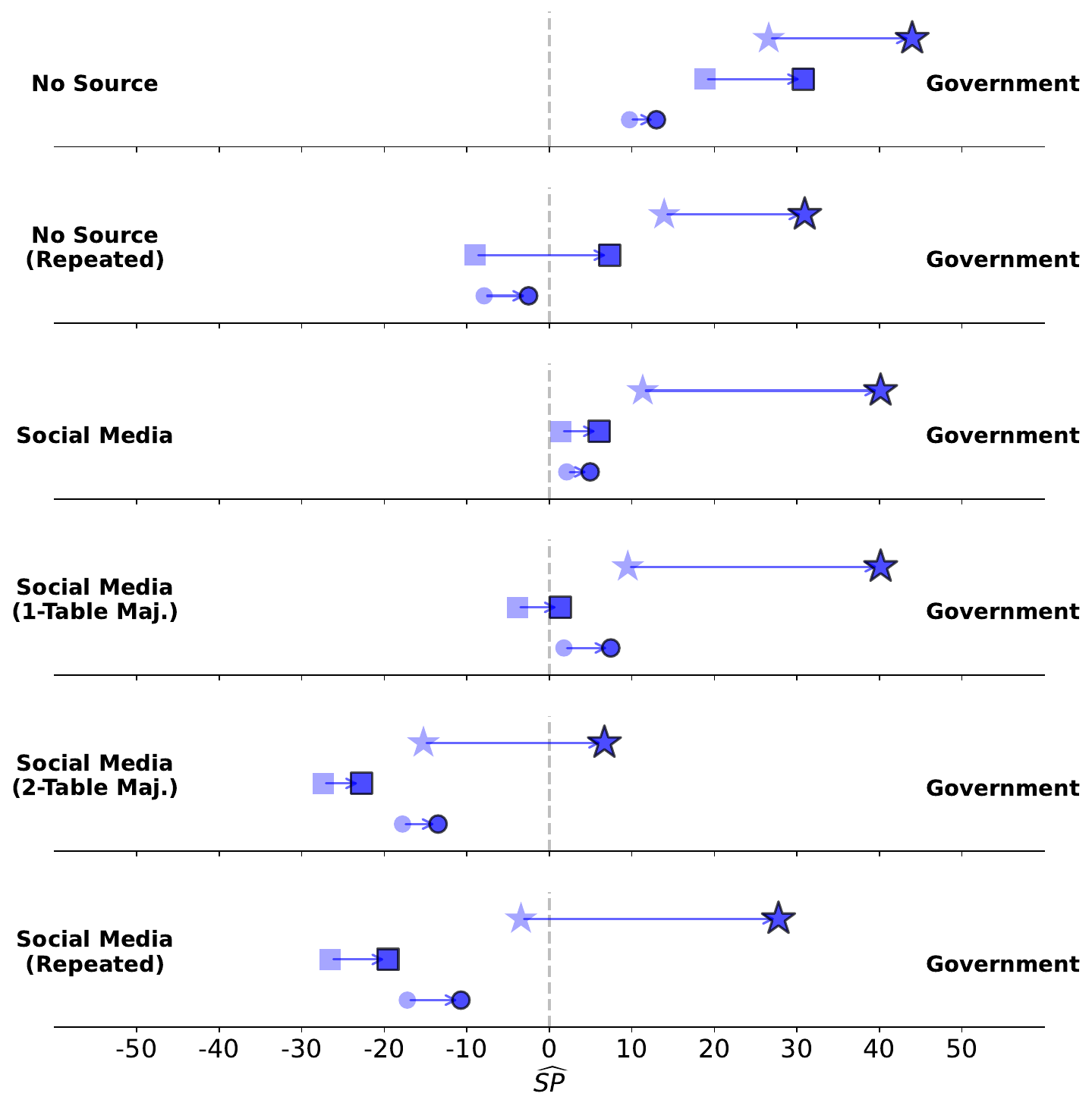}
    \caption{Source preferences  when \textsc{Llama} models are instructed to consider source credibility and repetition (darker), compared to original prompts (lighter). 
    This weakens repetition bias but not enough to ensure consistency with the original source hierarchy for all but the largest model. Legend in \S\ref{sec:setup-evaluation}.}
    \label{fig:promptbased_llama_rep}
\end{figure}

\begin{figure}[t]
    \centering
    \includegraphics[width=\linewidth]{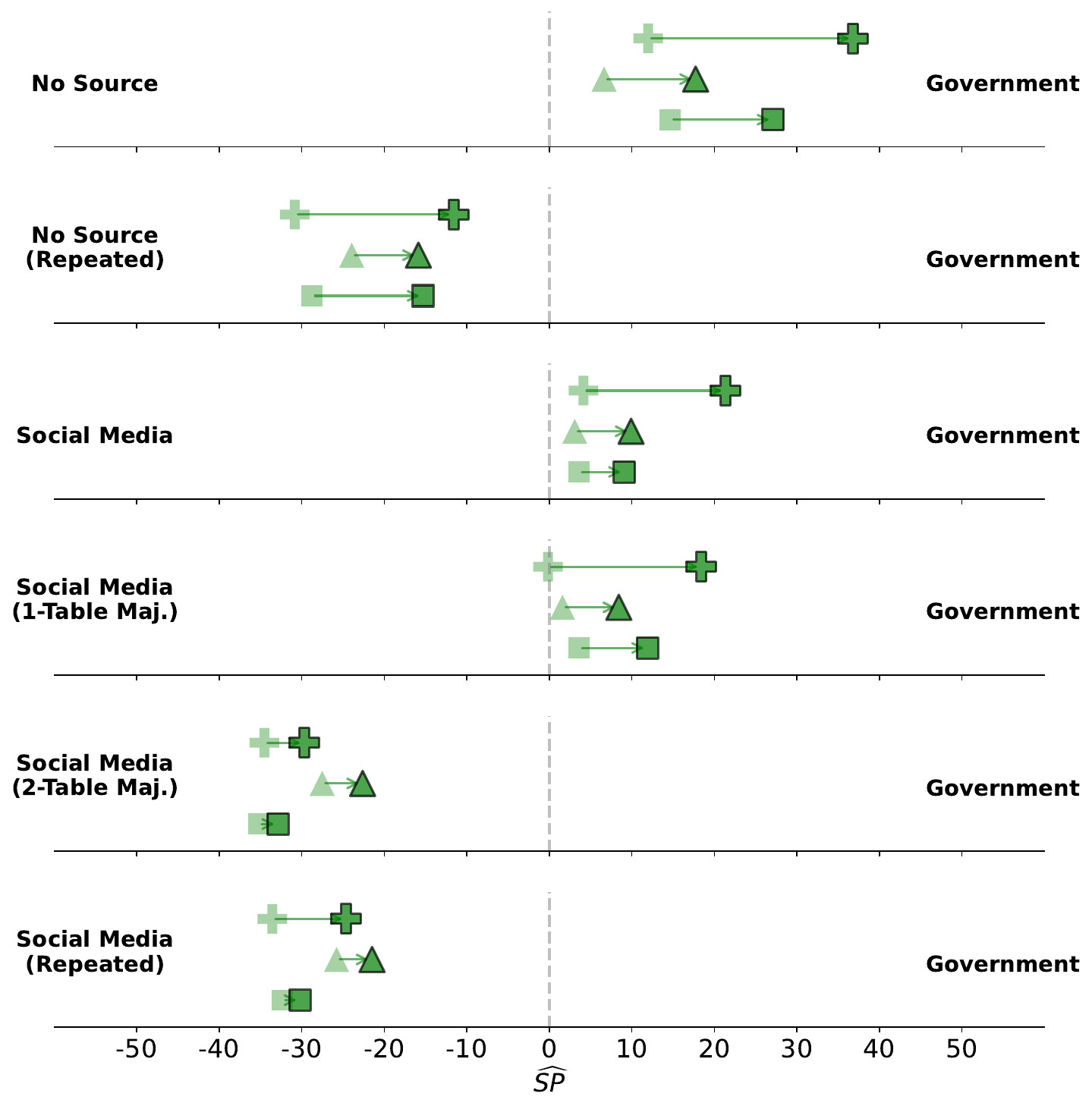}
     \caption{Source preferences  when \textsc{OLMo} models are instructed to consider source credibility and repetition (darker), compared to original prompts (lighter). 
    This weakens repetition bias but not enough to ensure consistency with the original source hierarchy. Legend in \S\ref{sec:setup-evaluation}.}
    \label{fig:promptbased_olmo_rep}
\end{figure}

\section{Repetition Prompting for All Models}
\label{sec:full_prompt_based_repetition}

We add a paragraph to the \textit{instruction} of every prompt, stating: \textit{"When selecting an answer, identify which sources support each option and assess the credibility of those sources before deciding. Simple repetition of information should not influence your choice."}.
Figures~\ref{fig:promptbased_gemma_rep}, \ref{fig:promptbased_llama_rep} and \ref{fig:promptbased_olmo_rep} show the impact of this mitigation strategy on the \textsc{Gemma}, \textsc{Llama} and \textsc{OLMo}  model families, respectively. 
The rightward shifts show that repetition prompting does strengthen original source preferences, partially with a greater effect at mitigating repetition bias compared to a true majority bias.
However, prompting is insufficient to ensure consistency with the source hierarchy in the absence of repetition, with the exception of the largest \textsc{Llama} model.

\begin{figure}[t]
    \centering
    \includegraphics[width=\linewidth]{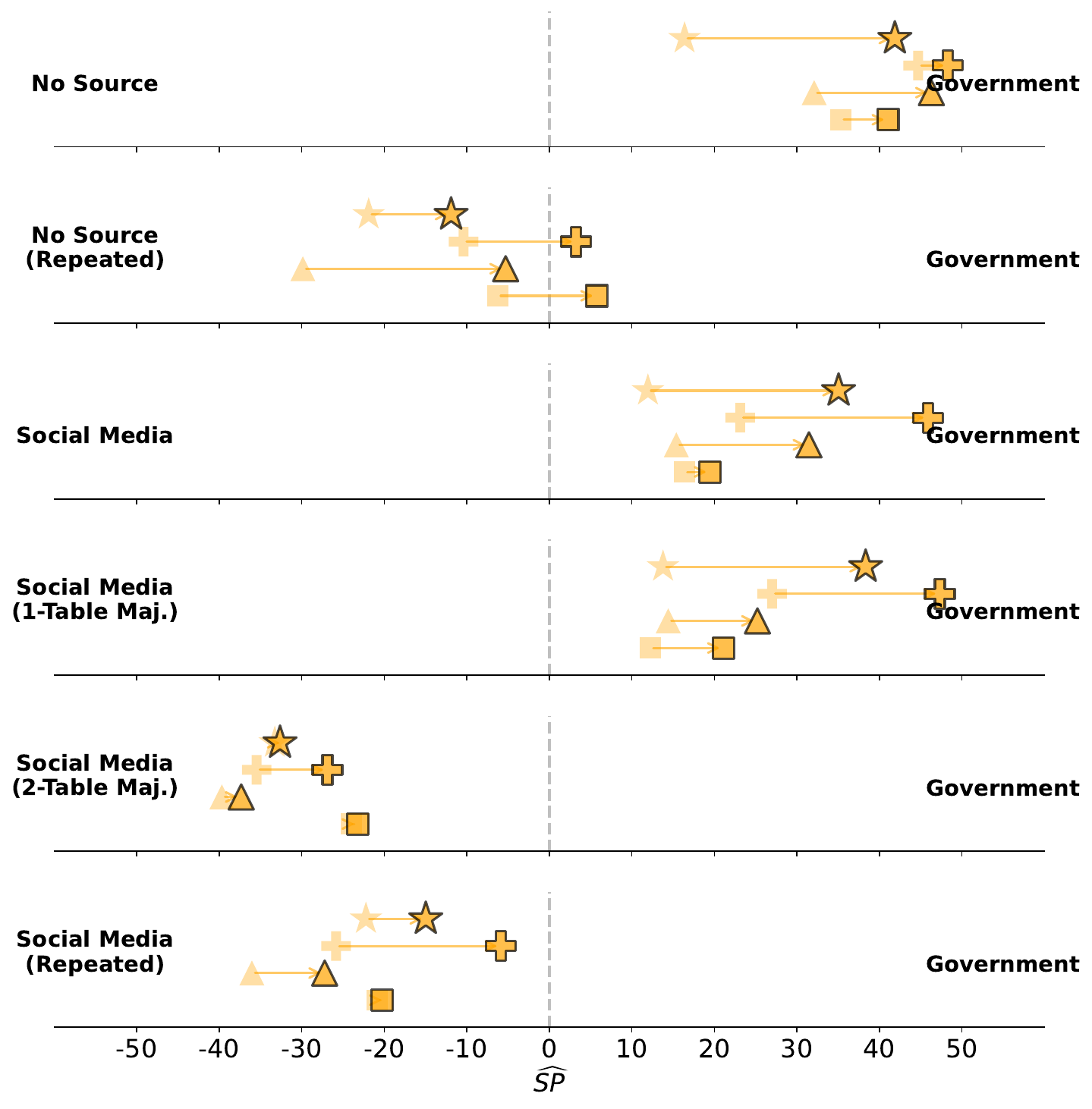}
    \caption{Source preferences  when \textsc{Qwen} models are instructed to consider source credibility (darker), compared to original prompts (lighter). 
    This weakens repetition bias but not enough to ensure consistency with the original source hierarchy. Legend in \S\ref{sec:setup-evaluation}.}
    \label{fig:promptbased_qwen_cred}
\end{figure}

\begin{figure}[t]
    \centering
    \includegraphics[width=\linewidth]{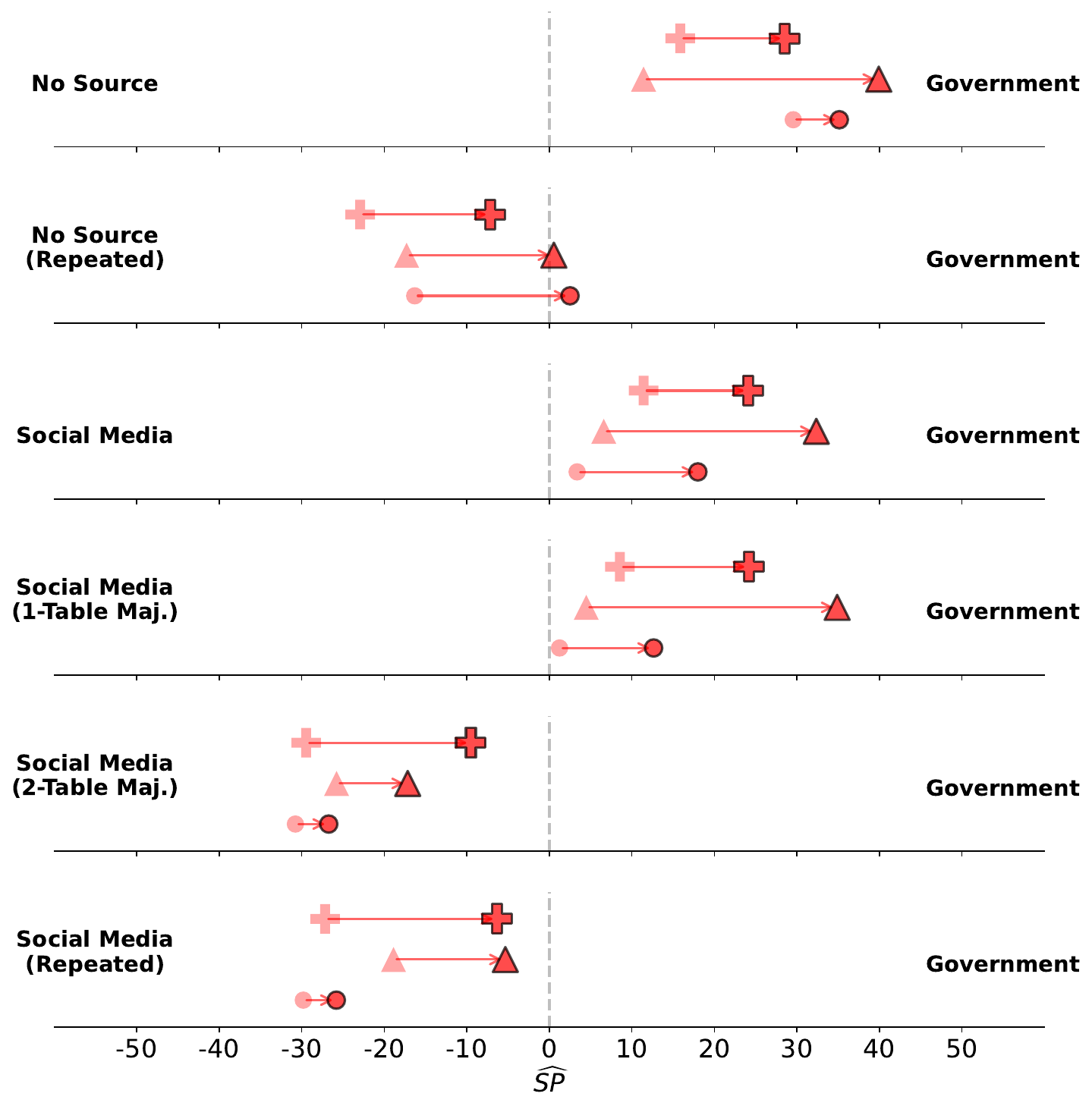}
    \caption{Source preferences  when \textsc{Gemma} models are instructed to consider source credibility (darker), compared to original prompts (lighter). 
    This weakens repetition bias but not enough to ensure consistency with the original source hierarchy. Legend in \S\ref{sec:setup-evaluation}.}
    \label{fig:promptbased_gemma_cred}
\end{figure}

\begin{figure}[t]
    \centering
    \includegraphics[width=\linewidth]{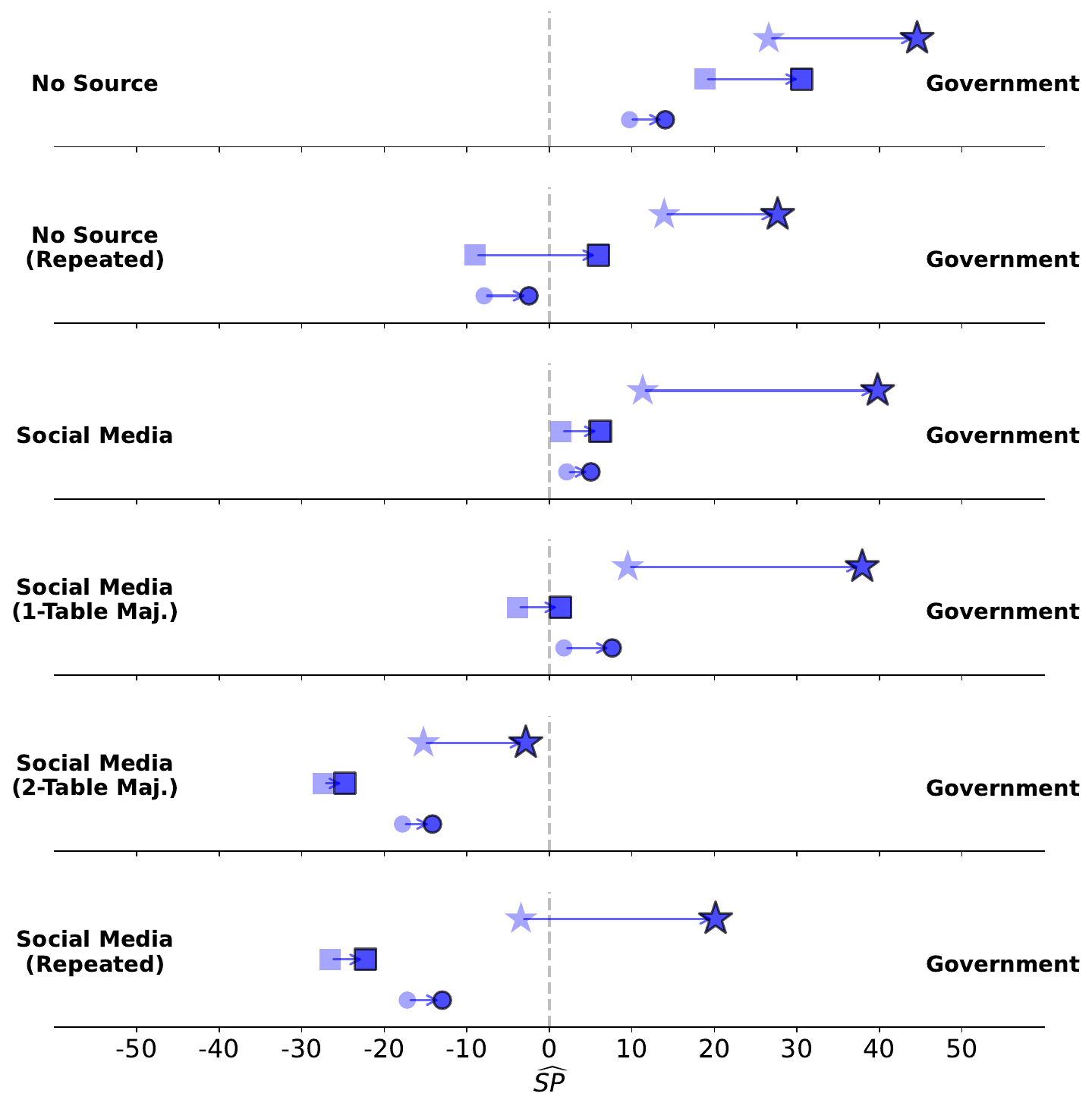}
    \caption{Source preferences  when \textsc{Llama} models are instructed to consider source credibility (darker), compared to original prompts (lighter). 
    This weakens repetition bias but not enough to ensure consistency with the original source hierarchy for all but the largest model. Legend in \S\ref{sec:setup-evaluation}.}
    \label{fig:promptbased_llama_cred}
\end{figure}

\begin{figure}[t]
    \centering
    \includegraphics[width=\linewidth]{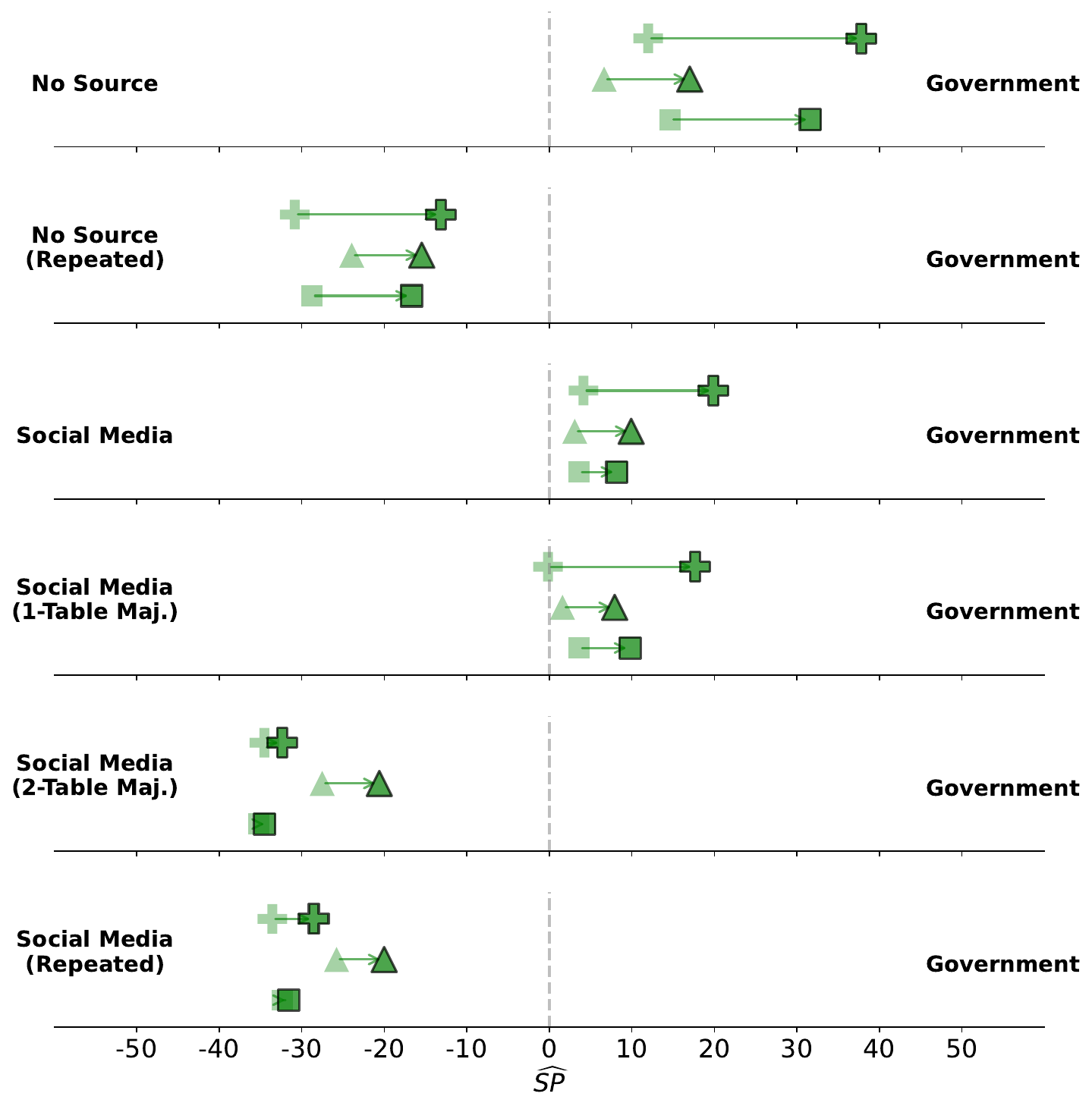}
     \caption{Source preferences  when \textsc{OLMo} models are instructed to consider source credibility (darker), compared to original prompts (lighter). 
    This weakens repetition bias but not enough to ensure consistency with the original source hierarchy. Legend in \S\ref{sec:setup-evaluation}.}
    \label{fig:promptbased_olmo_cred}
\end{figure}

\section{Credibility Prompting for All Models}
\label{sec:full_prompt_based_cred}

We add a paragraph to the \textit{instruction} of every prompt, stating: \textit{"When selecting an answer, identify which sources support each option and assess the credibility of those sources before deciding."}.
Figures~\ref{fig:promptbased_qwen_cred}, \ref{fig:promptbased_gemma_cred}, \ref{fig:promptbased_llama_cred} and \ref{fig:promptbased_olmo_cred} show the impact of this mitigation strategy on model families, which is similar but slightly weaker than the repetition-aware prompt.

\clearpage

\section{Fine-tuning-based Mitigation: Details}
\label{sec:mitigation-details}

\paragraph{Training data.}
Our training data knowledge conflicts are created from $12$ seed entities from \textsc{NeoQA} with all their original and counterfactually-perturbed attribute values.
To increase the number of conflict pairs, we also include conflicts between our counterfactually-perturbed entities, in contrast with all of our other evaluations where we always compare a \textsc{NeoQA} seed entity to a perturbed one.
Next, we add sources of different types to this data.
To test the generalization to unseen source types, we only include newspaper and person sources.
We combine sources and conflict pairs randomly to get a total of $1,500$ unique inputs of conflict pair and source match-ups, with $1,275$ used for training and $225$ used for validation.

\paragraph{Test data.}
Our test data consists of knowledge conflicts with all $361$ remaining seed entities from \textsc{NeoQA} and all their augmented versions.
These are paired exclusively with government sources and social media sources that have no overlap with the training data.

\paragraph{Experimental details.}

We use one Quadro RTX 6000 Nvidia GPU to fine-tune the \textsc{Gemma-3-4B} model. 
We use a batch size of $32$, a learning rate of $1e^{-4}$ with a warm-up period in the first $10\%$ of training steps.
We insert and fine-tune $16$ million LoRA parameters with typical hyperparameters of $r=8$, $\alpha =16$ and dropout of $0.05$.
$\lambda$ is set to $200$.
Every $32$ training steps, we evaluate the model on a small held-out validation split of $225$ conflict pairs.
Based on this validation, we use early stopping with a patience of $2$ to avoid overfitting for a maximum of $3$ epochs, where training terminates during the second epoch.
This is followed up by a $200$ step training epoch, only using the first KL-divergence loss term ($\lambda = 1$).
All hyperparameters were tuned to minimize $\widehat{SP}$ gaps and preserve the original source preferences. 
The total training time for one seed on our hardware is less than one hour.

\section{Hardware Specifications}

For models with more than $14$B parameters, we use 1-2 Nvidia H200 GPUs.
For smaller models, we use Nvidia Quadro RTX 6000 GPUs.
It takes up to six hours to run experiments with repeated tables and the largest model, \textsc{Llama-3.1-70B}.

\section{AI Assistance}

We used ChatGPT-5 for finding related work.
For coding, we used GitHub Copilot and Claude Code to refactor and document code, and to write boilerplate code for logging.
No AI assistance was used for writing or research design.

\end{document}